\documentclass{article}

\usepackage{arxiv,times}
\usepackage[british]{babel}

\usepackage{amsmath,amssymb}
\usepackage{array}
\usepackage{booktabs}
\usepackage{graphicx}
\usepackage{longtable}
\makeatletter
\let\LT@makecaption@body\LT@makecaption
\renewcommand{\LT@makecaption}[3]{\LT@makecaption@body{#1}{\normalsize#2}{#3}}
\makeatother
\usepackage{microtype}
\usepackage{placeins}
\usepackage{xurl}
\usepackage{xcolor}
\usepackage{hyperref}

\hypersetup{
  colorlinks=true,
  linkcolor=black,
  citecolor=blue,
  urlcolor=blue,
  pdftitle={TabBench-Bio: A Living Benchmark for Machine Learning on High-Dimensional Biomedical Tables},
  pdfauthor={Jules Kreuer, Sofiane Ouaari, Julia Hellmig, Julius Braitinger, Nico Pfeifer}
}

\graphicspath{{figures/}}
\newcommand{\ProjectName}{TabBench-Bio}

\newcommand{\BenchmarkDatasetCount}{43}
\newcommand{\ClassificationDatasetCount}{30}
\newcommand{\RegressionDatasetCount}{13}

\newcommand{\PredictivePeerModelCount}{19}

\newcommand{\TabularFoundationModelCount}{9}
\newcommand{\ModalityCount}{7}

\newcommand{\GridCellCount}{28}
\newcommand{\FeatureBudgetCount}{4}
\newcommand{\SampleBudgetCount}{7}
\newcommand{\FeatureBudgetSet}{\{2{,}000, 10{,}000, 25{,}000, \mathrm{full}\}}
\newcommand{\SampleBudgetSet}{\{20, 50, 100, 200, 500, 1{,}000, \mathrm{full}\}}
\newcommand{\CVFoldCount}{5}

\newcommand{\TimeLimitHours}{1}

\newcommand{\EloScale}{400}

\newcommand{\EloConfidenceLevel}{95}

\newcommand{\ReferenceCell}{p=10,000, n=100}
\newcommand{\ReferenceTargetCount}{42}

\newcommand{\AUROCTopModel}{RealTabPFN 2.5}
\newcommand{\AUROCSecondModel}{TabPFN Wide 5k (ne3)}
\newcommand{\AUROCThirdModel}{TabPFN 3}
\newcommand{\AUROCLeadingSpread}{20}

\newcommand{\AUROCTFMLeadCount}{6}
\newcommand{\PolicyTopModel}{RealTabPFN 2.5}
\newcommand{\PolicySecondModel}{TabICLv2}
\newcommand{\PolicyRankCorrelation}{0.72}
\newcommand{\PolicyMovedModelCount}{16}
\newcommand{\PolicyLargestMoveModel}{TabICLv2}
\newcommand{\PolicyLargestMoveRanks}{17}
\newcommand{\PolicyLargestMoveLostTargets}{31}

\newcommand{\ModalityLeaderMethylation}{TabPFN Wide (8k)}

\newcommand{\ModalityLeaderGenomicPrediction}{Random Forest}

\newcommand{\ModalityRunnerUpGenomicPrediction}{Extra Trees}

\newcommand{\ModalityTargetsGenomicPrediction}{4}

\newcommand{\ModalityTargetsMetagenomics}{10}

\newcommand{\ContrastTargetCount}{34}

\newcommand{\TabICLFeatureLimit}{2,000}

\newcommand{\TabICLFeatureRankNarrow}{17}
\newcommand{\TabICLFeatureRankWide}{19}

\newcommand{\LogisticRegressionFeatureRankNarrow}{7}
\newcommand{\LogisticRegressionFeatureRankWide}{4}

\newcommand{\TopModelElo}{1156}
\newcommand{\TopModelEloLow}{1098}
\newcommand{\TopModelEloHigh}{1212}

\newcommand{\RunnerUpElo}{1070}

\newcommand{\TopEloMargin}{86}
\newcommand{\TopPairEloDiff}{87}
\newcommand{\TopPairEloDiffLow}{43}
\newcommand{\TopPairEloDiffHigh}{129}

\newcommand{\TopPairWinRatePercent}{78.6}
\newcommand{\TopPairWinRateLowPercent}{65.5}
\newcommand{\TopPairWinRateHighPercent}{89.3}
\newcommand{\TopPairWinCount}{32}

\newcommand{\TopPairTargetCount}{42}
\newcommand{\BestMacroFModel}{Logistic Regression}
\newcommand{\BestMacroFValue}{0.721}
\newcommand{\BestTFMModel}{RealTabPFN 2.5}
\newcommand{\BestTFMMacroF}{0.713}

\newcommand{\TopModelMedianFitSeconds}{78.9}
\newcommand{\FastCompetitiveModel}{Logistic Regression}
\newcommand{\FastCompetitiveFitSeconds}{33.1}

\title{\ProjectName: A Living Benchmark\\
for Machine Learning on High-Dimensional\\
Biomedical Tables}

\author{%
Jules Kreuer \quad Sofiane Ouaari \quad Julia Hellmig \quad Julius Braitinger \quad Nico Pfeifer\\[0.6em]
\small Methods in Medical Informatics, University of Tübingen, Tübingen, Germany}
\date{}

\begin{document}
\maketitle

\begin{abstract}
Biomedical tables often combine thousands of measured variables with only tens or hundreds of labelled samples, a regime that is poorly represented in general-purpose tabular benchmarks. We introduce TabBench-Bio, a living and interactive benchmark of \BenchmarkDatasetCount{} biomedical datasets spanning multiple domains. Under a shared cross-validation protocol, we compare classical estimators, neural networks, and tabular foundation models across \GridCellCount{} feature-by-sample operating points.
At the reference cell of 10,000 features and 100 training samples, RealTabPFN 2.5 has the highest point estimate, closely followed by TabPFN 3 and Logistic Regression. A paired bootstrap over the target pool separates RealTabPFN 2.5 from TabPFN 3 by \TopPairEloDiff{} Elo (\EloConfidenceLevel\% interval [\TopPairEloDiffLow{}, \TopPairEloDiffHigh{}]). Tabular foundation models generally occupy the leading ranks, while the strongest configuration depends on the operating point and biomedical modality. The AutoML framework AutoGluon, using its one-hour ``extreme'' preset, is configured as a separate resource-intensive reference and reported here at the reference and full cell. Fold-level predictions, run status, and deterministic aggregations make every reported result reproducible and reusable.
The benchmark is open to contributions of new biomedical datasets.
The interactive leaderboard is available at
\url{https://tabbench-bio.eu}.

\end{abstract}

\section{Introduction}

Comparative evaluation of machine learning on biomedical tables remains heterogeneous, as individual studies typically select limited cohorts, assays, and estimators. Differences in preprocessing and validation make results difficult to compare, a problem exacerbated by the arrival of tabular foundation models (TFMs) \citep{pfn}, whose priors may help when data is scarce but whose computational behaviour differs from linear models and boosted trees. Furthermore, most comparisons report only one operating point per dataset, obscuring whether a model is robust to feature width or cohort size in high-dimensional, low-sample-size (HDLSS) settings. We therefore treat a controlled feature-by-sample grid as a primary part of the benchmark definition.

General tabular benchmarks like TabArena \citep{tabarena} and TALENT \citep{talent} provide a foundation by curating datasets, implementations, and leaderboards, while \citet{grinsztajn2022} identify when tree-based models remain strong. However, these focus primarily on conventional independent and identically distributed (IID) tasks and provide limited evidence on the HDLSS geometries common in molecular biomedicine.

In molecular biomedicine, gene-expression, metagenomic, and methylation profiles often expose thousands of variables for only tens or hundreds of samples. In this HDLSS regime, the feature-to-sample ratio affects both statistical estimation and model feasibility. Classical estimators face irrelevant measurements, neural networks lack sufficient supervision, and TFMs may exceed their original feature limits. For instance, TabPFN v2's performance on small-to-medium tables \citep{tabpfnv2} does not establish how its prior behaves when feature width and cohort size are varied independently.

RamanBench \citep{ramanbench} highlights the value of domain-focused HDLSS benchmarks. Raman spectra have a physically ordered wavenumber axis, whereas the gene-expression and metagenomic tables considered here do not share an analogous column ordering. Biomedical HDLSS data therefore require a benchmark spanning multiple measurement processes rather than treating one assay as a proxy for the whole regime.

We introduce TabBench-Bio, a living benchmark combining a versioned registry, common execution interface, and deterministic reporting. Following TabArena's maintained perspective and RamanBench's domain-specific motivation, we evaluate a \FeatureBudgetCount-by-\SampleBudgetCount{} grid (\GridCellCount{} operating points) with shared cross-validation folds and nested feature caps. This ensures the held-out population remains fixed, making robustness across data geometries a primary outcome. Our main contributions are as follows:

\begin{itemize}
  \item a versioned registry of \BenchmarkDatasetCount{} biomedical datasets spanning
  heterogeneous molecular and sequence-derived representations, all evaluated under the
  shared benchmark protocol;
  \item a common implementation-level comparison of \PredictivePeerModelCount{}
  predictive model configurations and a constant baseline, with AutoGluon configured as a
  separate resource-intensive AutoML reference at two prespecified operating points;
  \item a leak-resistant \FeatureBudgetCount-by-\SampleBudgetCount{} grid with
  fold-aligned evaluation, explicit failure semantics, and cell-conditional
  Bradley-Terry rankings;
  \item reusable fold-level predictions, run status,
  resource measurements, deterministic aggregations, and an interactive leaderboard.
\end{itemize}

\section{Related work}

Earlier tabular studies established that the relative performance of trees and neural networks depends on dataset scale, feature type, and tuning protocol \citep{grinsztajn2022,somvanshi2026survey}. TALENT \citep{talent} and TabArena \citep{tabarena} provide frameworks for consistent evaluation and benchmark governance, with TabArena emphasizing a living leaderboard and reusable predictions. TabBench-Bio adopts this versioned structure but focuses on biomedical HDLSS tasks and the robustness of fixed configurations across resource regimes, rather than extensively tuned performance. This domain shift is critical: TabArena's findings on foundation models and tuning \citep{tabarena} describe general-purpose tasks and may not carry over to biomedical HDLSS data.

Gradient-boosted trees (XGBoost, LightGBM, CatBoost) remain essential baselines for heterogeneous tables \citep{xgboost,lightgbm,catboost}. Neural tabular learning has evolved from generic MLPs to specialised architectures like RealMLP \citep{realmlp} and TabM \citep{tabm}. Tabular foundation models (TFMs) for in-context learning include RealTabPFN v2, 2.5 for small-data prediction, TabPFN-Wide for high feature counts, TabPFN 3, TabDPT, and Mitra \citep{realtabpfn,tabpfnv25,tabpfnv3,tabpfnwide,mitra,tabdpt}, as well as TabICLv2 \citep{tabiclv2}, the successor to TabICL \citep{tabicl}, and TabFM \citep{tabfm}. Because these models make different assumptions about how evidence scales with rows and columns, TabBench-Bio's two-dimensional grid makes these dependencies visible.

RamanBench \citep{ramanbench} provides a methodological precedent through its live leaderboard for Raman spectroscopy. Architectures using spectral locality require an appropriate representation or adaptation before application to omics tables without an analogous column ordering. TabBench-Bio examines whether general-purpose tabular learners persist across heterogeneous biomedical representations that share no physical column axis.

Various resources provide biomedical data: OpenML \citep{openml}, GEO and TCGA for molecular profiles \citep{geo,tcga}, MGnify for microbiome studies \citep{mgnify}, TDC for drug discovery \citep{tdc}, and ChEMBL for bioactivity \citep{chembl2023}. Genomic-prediction panels also link SNPs to quantitative traits \citep{azodi2019}, while protein-language-model benchmarks provide sequence-derived representations \citep{esm2,deeploc2,peer}. TabBench-Bio unifies these sources in one registry, evaluating them under a shared protocol for splitting, fitting, scoring, and aggregation.

\section{TabBench-Bio}
\label{sec:benchmark}

TabBench-Bio evaluates supervised classification and regression on static, pre-extracted numerical representations of biomedical samples in the HDLSS regime.

Using default configurations and fixed validation folds, we ask how tabular learners behave as the number of labelled samples and the feature width change. Our current scope excludes survival analysis, longitudinal modelling, multimodal integration, external-site validation, and supervised feature selection. Each would change the prediction unit, endpoint, or source of supervision addressed by the current protocol.

\subsection{Dataset scope and selection}

The registry contains \BenchmarkDatasetCount{} datasets: \ClassificationDatasetCount{} classification and \RegressionDatasetCount{} regression tasks. MGnify and OpenML contribute nine and ten datasets, respectively; the remainder come from TCGA, GEO, TDC, ChEMBL, Dryad/Zenodo, and curated or locally derived releases. The datasets span \ModalityCount{} modality groups with different measurement processes, sparsity, and dependence structures, allowing comparisons across biological representations rather than a single assay. Appendix~\ref{app:datasets} lists the full registry.

The collection includes six molecular-property tasks from TDC, two single-protein pChEMBL regression endpoints from ChEMBL 37, four crop genomic-prediction tasks, four DNA methylation cohorts from GEO, two protein-embedding tasks, and a metagenomic marker-profile cohort for liver cirrhosis. For the ChEMBL tasks, we use exact, unflagged activity measurements \citep{chembl2023}, ECFP4 fingerprints \citep{rogers2010}, and Bemis-Murcko scaffolds as split groups \citep{bemis1996}. The crop tasks pair SNP panels for maize, rice, soybean, and switchgrass with flowering-time, yield, or height phenotypes \citep{azodi2019}; the methylation tasks retain their source-study endpoints \citep{hannum2013,liu2013}. Three locally constructed datasets use mean-pooled embeddings: Evo 2 for BRCA1 variant sequences and ESM-2 for DeepLoc 2.0 fungal proteins and PEER TEM-1 mutants \citep{evo2,findlay2018,esm2,deeploc2,peer}. They use biological locus, published homology partition, and mutation position, respectively, as cross-validation groups (Appendix~\ref{app:local-construction}).

We combined automated shortlisting with manual curation. Automated preselection generally required at least 70 samples and 1,000 features, non-synthetic data, and an open or reusable licence. We tested these candidates for predictive signal using a random forest with 100 label permutations, requiring \(p<0.05\) and a cross-validated score of at least \(0.10\) adjusted balanced accuracy for classification or \(0.05\) \(R^2\) for regression; we excluded near-perfect scores above \(0.98\) as trivial. These low minimum thresholds were intended to exclude candidates with little detectable signal, rather than require strong RF performance. The scan provided a preliminary shortlist; manual additions were selected for relevance and quality. We checked the remaining candidates for duplicate observations, missing or constant features, inconsistent representations, outliers, potential target leakage, batch-target association, and class balance. Final review covered each target, prediction unit, licence, group identifiers, and reproducible table construction. The registry retains stable identifiers and the reviewed metadata. Before constructing folds, we remove classes with fewer than ten samples, so the benchmark does not assess rare-class recognition.

\begin{figure}[t]
  \centering
  \includegraphics[width=0.95\linewidth]{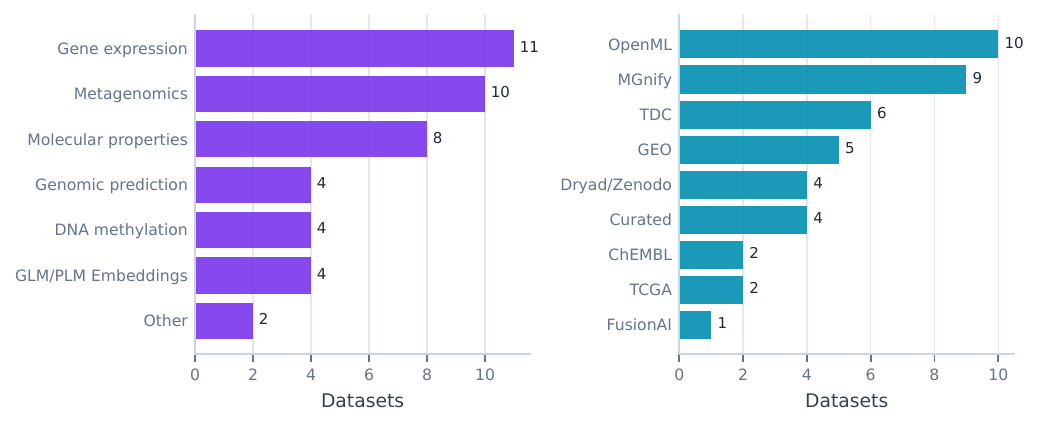}
  \caption{Benchmark dataset counts by modality group and source, generated from the
  versioned registry.}
  \label{fig:composition}
\end{figure}

\subsection{Models and fitting regime}

The peer comparison comprises \PredictivePeerModelCount{} predictive configurations:
L2-regularised Logistic Regression for classification and Ridge Regression for regression; nearest neighbours; random forests and extra trees; CatBoost, LightGBM, and XGBoost; a standard MLP, RealMLP, and TabM; and
\TabularFoundationModelCount{} TFMs, including RealTabPFN v2, 2.5, TabPFN 3, two TabPFN-Wide configurations, Mitra, TabDPT, TabICLv2, and TabFM \citep{realtabpfn,tabpfnv25,tabpfnv3, tabpfnwide,mitra,tabdpt,tabiclv2,tabfm}. We fit a constant predictor as an explicit baseline and use its score to impute out-of-memory or time-limit failures. The benchmark exposes all models through a common adapter built on the AutoGluon tabular interface \citep{autogluon}.

For each peer, we use the default configuration exposed by the registered AutoGluon model adapter, without per-dataset hyperparameter search. Benchmark-level bagging, stacking, and ensembling are disabled. Results concern these fixed configurations; the effects of bagging and hyperparameter tuning are not evaluated.

Appendix~\ref{app:model-defaults} lists the principal configuration parameters for every model; values not listed remain at the corresponding library defaults.

The benchmark measures how reliable each model is as a reusable default rather than the best performance
obtainable after bespoke tuning. Fitting and prediction are each subject to the same external timeout of \TimeLimitHours{} hour.
Each model fit was restricted to 32 threads. All GPU-based models, including the native AutoGluon Extreme reference, were evaluated on NVIDIA L40S GPUs with 48 GB of device memory, with one fit per GPU. CPU-only models were evaluated on CPU nodes. All models within a cell used the same frozen data splits and model configurations. GPU type and available memory were recorded with each run. The reported results therefore reflect the specified memory and wall-clock budgets.
When prediction exhausted GPU memory, we retried the same held-out fold in smaller query batches, starting with the full fold and repeatedly halving the batch down to one row if necessary. Classification labels and probabilities were handled independently and concatenated in the original row order. The fitted model and held-out rows were retained, and the original timeout continued to apply. Runs that exhausted memory after the applicable retries were recorded as failures.
We evaluated AutoGluon's native Extreme preset at two prespecified points: the 10,000-feature, 100-sample reference cell and the full-feature, full-sample cell (Figure~\ref{fig:elo-full}). Its one-hour budget made evaluation throughout the grid computationally prohibitive. The results below use the reference cell. We report it as a resource-intensive one-hour AutoML reference rather than a ranked peer.
\subsection{Operating-point grid}

The grid evaluates combinations of feature width and labelled training-sample count. Its settings are
\[
  p \in \FeatureBudgetSet, \qquad
  n \in \SampleBudgetSet,
\]
leading to 28 different combinations. The prespecified reference cell uses 10,000 features and 100 training samples. Within each cell, all models share the same folds and training rows; the evaluation unit is a fold-target-model combination.

Sample budgeting affects only the training partition; all budgets share the same held-out fold. Regression uses seeded random subsampling, whereas classification allocates the requested budget across classes and samples within each class using the same seed.
The number of training rows selected from each class is recalculated at each budget, thus a larger training subset may therefore omit rows included in a smaller subset. Data-dependent feature filtering is fitted separately on each budgeted training set. Sample-budget comparisons therefore characterize the complete budgeted learning procedure rather than isolating sample count while holding all other components fixed.

We distinguish a cell's \emph{nominal} sample budget from the \emph{effective} count used after a memory failure. In strict mode, successful runs contribute their metrics at the requested nominal cell, attempted failures follow the prespecified failure policy, and design exclusions remain absent; no result from a smaller sample-budget cell is substituted. If fitting, or prediction after query batching, exhausts memory, the adaptive sensitivity mode searches smaller configured budgets for the same dataset, target, fold, model, and feature cap, approximately halving the count until a completed result is found.

We vary feature width to assess how predictive performance and computational feasibility change with the number of available inputs. For gene-expression, methylation and genomic-prediction datasets, smaller subsets approximate settings where fewer genes, CpG sites or variants are measured. For metagenomic profiles, fingerprints and embeddings, they test sensitivity to retaining only part of the available representation. Random subsets do not reproduce a particular assay or a biologically selected panel.

Using only the budgeted training data, we remove features that are missing in the majority of rows. Missingness filtering may discard predictive information, and source-level curation does not guarantee nontriviality at every subsampled operating point. For each fold, we place the remaining columns in a seeded random order. Smaller feature caps are prefixes of this ordering, and we restore selected columns to registry order before fitting. Feature sets are thus nested across caps and identical across models within a fold and sample budget; each of the \CVFoldCount{} folds uses a different ordering. Because missingness filtering is refitted after sample budgeting, the eligible feature set can differ across sample budgets. Increasing the retained width changes both dimensionality and the information available to the model. The feature sweep therefore does not isolate dimensionality alone or evaluate supervised feature selection. At the reference cell, 28 of 42 datasets retain 10,000 features in all five folds; the others have fewer available features (Appendix~\ref{app:sample-composition}).

Missing values are passed unchanged to models that handle them natively. For nearest neighbours, missing values are imputed with medians estimated from the training data. No transformation, allocation decision, or feature filter consults the held-out fold. Models that support only classification, or only regression, are evaluated on those tasks alone.

\subsection{Cross-validation and leakage control}

We use \CVFoldCount-fold cross-validation, with complementary folds shared across all cells and models. Sample caps restrict training data only. A single partition does not characterize sensitivity to alternative splits, particularly with few independent groups. When a loader supplies biological group
identifiers, classification uses stratified group folds and regression uses group folds, so no group crosses the train-test boundary. If such identifiers are unavailable, classification uses stratified row-wise folds and regression uses row-wise folds. Without reliable group metadata, row-wise evaluation cannot establish patient- or site-level independence; duplicate screening does not exclude hidden biological dependence. The registry records the split unit explicitly; we do not infer it from apparent similarities among measurements.
\subsection{Outcomes and aggregation}

Macro-F1 is the primary classification metric, weighting classes equally without designating a positive class; RMSE is the primary regression metric. We save fold-level predictions and ground truth, then recompute the metrics during aggregation.

Within each cross-validation fold of each target, we compare every pair of eligible models using observed or
failure-imputed scores. Because all comparisons are made within targets, classification and regression metrics are never placed on the same arithmetic scale. Under the Bradley-Terry model, the expected comparison score of model \(i\) against model \(j\) is
\begin{equation}
  E_{ij} = \frac{1}{1 + {10}^{(R_j-R_i)/400}},
  \qquad
  R_i = 1000 + 400\,(\theta_i - \theta_{\mathrm{RF}}).
\end{equation}
Wins, ties and losses are scored as 1, 0.5 and 0. Here \(\theta_i\) is the latent ability of model \(i\) in base-10 log-odds units and \(R_i\) is its fitted Elo rating, therefore a difference of \EloScale{} rating points corresponds to 10:1 odds. Each fold comparison is weighted by the reciprocal of the target's number of folds, so every target contributes the same total weight regardless of endpoint type, while a model that wins three of five folds is distinguished from one that wins all five. For interpretability, we shift the ratings so that Random Forest has an Elo of 1,000. We resample the target pool 2,000 times and report the rounded median of those draws as the rating, with \EloConfidenceLevel\% percentile intervals from the same draws. We resample targets rather than folds, carrying each target's fold comparisons with it, because folds are repeated measurements of the same scientific task; the target-bootstrap intervals therefore do not fully capture split uncertainty. We score classification on macro-F1 because the shared adapter also uses it to score the internal validation split, so it governs validation-based model selection and early stopping wherever a model exposes them. Models fitted by their own criterion, including the linear, nearest-neighbour, and tree baselines, are scored on macro-F1 without being optimised for it, so the alignment between fitting objective and reported metric is not uniform across the roster. We also report mean macro-F1 and median fit time, the latter over successful fits alone, so cost is conditional on a completed fit while performance charges a failure at constant-predictor level; we use Elo for cross-task ranking because it depends only on within-target comparisons \citep{bradleyterry}. Decision-threshold handling follows the AutoGluon adapter; Appendix~\ref{app:metric-sensitivity} re-ranks the reference cell by AUROC. The anchor shift is applied inside every bootstrap draw, so each reported interval is an interval for that model's rating \emph{relative to Random Forest}; the anchor's own interval is therefore degenerate. Because two such intervals share the anchor term, they are positively correlated and their overlap is not a test of one model against another. For a pairwise claim we instead take the difference between the two ratings within each draw, where the anchor cancels. A cell is designated complete when every scheduled unit in that cell has a terminal status, and the release manifest records the contributing cells and inputs. The reference-cell ranking in Figure~\ref{fig:elo} and Table~\ref{tab:leaderboard} uses the strict nominal-cell export. The strict mode is also the default view on the interactive website, which additionally provides the adaptive sensitivity view and rankings under balanced accuracy, Matthews correlation, and AUROC. Both exports and the sample-fallback manifest are released.

\section{Results}

The benchmark registry holds \BenchmarkDatasetCount{} datasets (\ClassificationDatasetCount{} classification and \RegressionDatasetCount{} regression), evaluated across \GridCellCount{} operating points. A dataset is included in a cell only if its training partition is at least as large as that cell's sample budget; otherwise, the fit would simply repeat the full-sample result. Consequently, the reference cell ranks \ReferenceTargetCount{} of them. TFMs generally occupy the leading ranks, and a paired bootstrap separates the leading model from its runner-up at the reference cell. Model order varies by operating point and biomedical modality, while fit time may favour a method other than the highest-rated one. We examine these patterns and their associated uncertainty below.
\subsection{Reference-cell ranking}

\begin{figure}[p]
  \centering
  \includegraphics[width=0.94\linewidth]{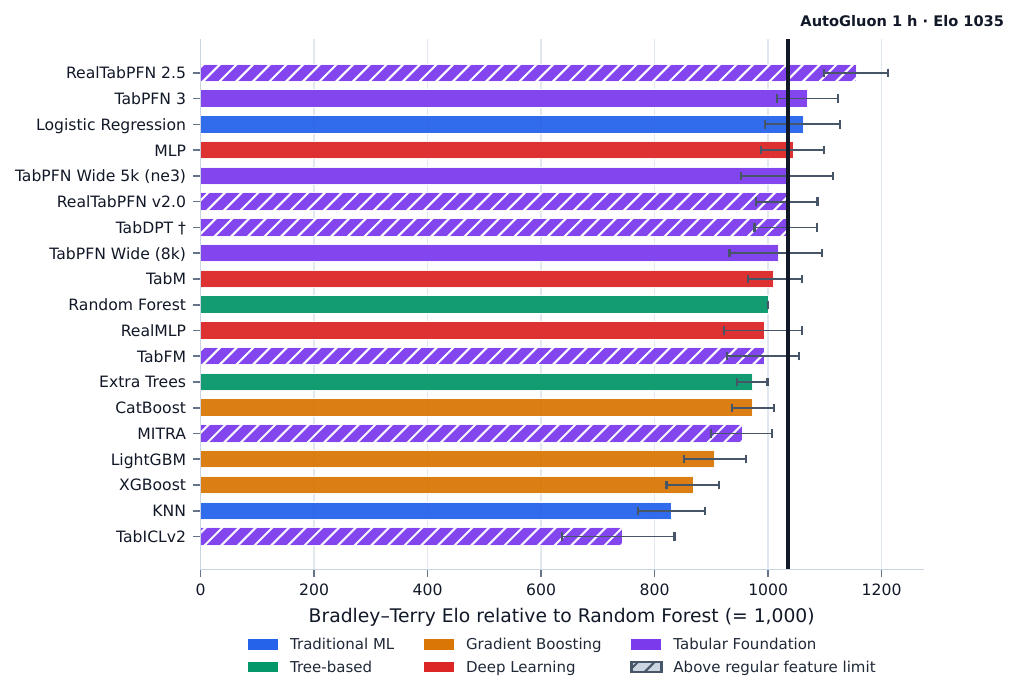}
  \caption{Strict reference-cell Elo at \ReferenceCell{}, anchored at Random Forest = 1,000. Colours denote model families; error bars show 95\% target-bootstrap intervals relative to this anchor. Interval overlap is not a pairwise test; pairwise claims use paired bootstrap differences. The vertical line marks the separate one-hour AutoGluon reference, not a ranked peer. \dag{}~TabDPT: see the pretraining-overlap caveat in Limitations.}
  \label{fig:elo}

  \vspace{0.1em}
  \includegraphics[width=0.94\linewidth]{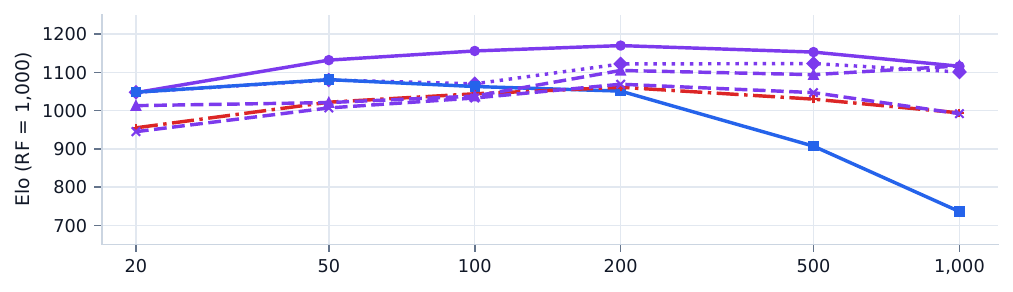}
  \par\vspace{0.1em}
  \includegraphics[width=0.94\linewidth]{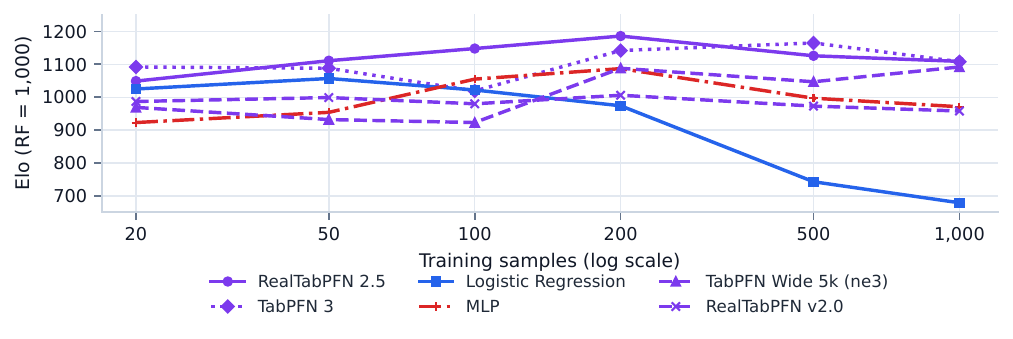}
  \caption{Sample-budget response at a 10,000-feature cap. Upper: all eligible datasets, whose composition varies across settings. Lower: the same Elo calculation restricted to the 14 datasets represented at all six settings. Lines show medians of 2,000 whole-target bootstrap draws for the six leading peers at the 100-sample reference cell; the lower legend applies to both panels. Strict failure scoring and model applicability are unchanged; TabPFN-Wide covers seven classification datasets in the lower panel.}
  \label{fig:sample-budget}
\end{figure}

The reference operating point is  defined with \ReferenceCell{}. We selected it before inspecting model results as it combines a high-dimensional feature budget with broad task coverage at a cohort size typical of the collection. At this point, the sample budget binds for \ReferenceTargetCount{} targets. The remaining dataset has fewer training rows than the cell's sample budget, so the cell would repeat their full-sample result and they do not enter the ranking. Figure~\ref{fig:elo} and Table~\ref{tab:leaderboard} summarise the strict nominal-cell results.

RealTabPFN 2.5 has the highest point estimate, with Elo \TopModelElo{} ([\TopModelEloLow{}, \TopModelEloHigh{}]). TabPFN 3 follows at \RunnerUpElo{}, a gap of \TopEloMargin{} points. Resampling the target pool and taking the rating difference within each draw gives \TopPairEloDiff{} Elo ([\TopPairEloDiffLow{}, \TopPairEloDiffHigh{}]), with RealTabPFN 2.5 rated above TabPFN 3 in all 2000 draws. The coarser target-level win count agrees. In their direct target-level comparison, RealTabPFN 2.5 wins \TopPairWinCount{} of \TopPairTargetCount{} targets, ties two, and loses in eight. Counting a tie as one half, its empirical win score is \TopPairWinRatePercent\% (target-bootstrap \EloConfidenceLevel\% interval [\TopPairWinRateLowPercent\%, \TopPairWinRateHighPercent\%]); the interval excludes 50\%.
Mean classification performance orders the leading models differently: \BestMacroFModel{} reaches the highest macro-F1, \BestMacroFValue{}, compared with \BestTFMMacroF{} for \BestTFMModel{}. The strongest TFM leads the rating, while a linear model remains a strong default in this biomedical HDLSS setting under macro-F1. Only two of the 30 classification tasks have median macro-F1 across configurations of at least \(0.95\) at the reference setting; Appendix~\ref{app:selection-difficulty} reports per-dataset difficulty.

As runs that exhaust memory or exceed the one-hour time limit are imputed with constant predictor performance, the benchmark indirectly ranks models by deployability. It balances predictive accuracy with the ability to operate within a fixed resource budget, penalizing models that fail to fit a target instead of omitting them from the results.

Failure counts differ substantially across models, so the two components should be read together; Appendix~\ref{app:failure-rates} reports recorded outcomes per model at this cell and Appendix~\ref{app:failure-policy} re-ranks the cell with failures omitted instead of imputed.

\begin{table}[t]
  \centering
  {\small
  \setlength{\tabcolsep}{4pt}
  \renewcommand{\arraystretch}{0.98}
  \begin{tabular}{@{}rlrrrrr@{}}
\toprule
Rank & Model & Elo & 95\% CI & Targets & Macro-F1 & RMSE $\downarrow$ \\
\midrule
1 & RealTabPFN 2.5 & 1156 & [1098, 1212] & 42 & 0.713 & 1.743 \\
2 & TabPFN 3 & 1070 & [1016, 1123] & 42 & 0.685 & 1.837 \\
3 & Logistic Regression & 1063 & [994, 1127] & 42 & 0.721 & 1.956 \\
4 & MLP & 1044 & [987, 1098] & 42 & 0.701 & 2.116 \\
5 & TabPFN Wide 5k (ne3) & 1038 & [952, 1114] & 30 & 0.675 & -- \\
6 & RealTabPFN v2.0 & 1033 & [979, 1087] & 42 & 0.665 & 1.818 \\
7 & TabDPT & 1032 & [976, 1086] & 42 & 0.694 & 2.095 \\
8 & TabPFN Wide (8k) & 1019 & [932, 1095] & 30 & 0.668 & -- \\
9 & TabM & 1010 & [964, 1060] & 42 & 0.681 & 2.015 \\
10 & Random Forest & 1000 & [1000, 1000] & 42 & 0.666 & 1.930 \\
11 & RealMLP & 993 & [922, 1060] & 42 & 0.699 & 2.171 \\
12 & TabFM & 993 & [927, 1055] & 42 & 0.633 & 1.791 \\
13 & Extra Trees & 973 & [945, 999] & 42 & 0.644 & 1.998 \\
14 & CatBoost & 973 & [937, 1011] & 42 & 0.645 & 1.978 \\
15 & MITRA & 955 & [900, 1007] & 42 & 0.647 & 1.903 \\
16 & LightGBM & 906 & [852, 961] & 42 & 0.639 & 1.881 \\
17 & XGBoost & 868 & [821, 914] & 42 & 0.635 & 2.110 \\
18 & KNN & 829 & [770, 889] & 42 & 0.598 & 2.220 \\
19 & TabICLv2 & 743 & [637, 835] & 42 & 0.360 & 2.168 \\
\bottomrule
\end{tabular}
}
  \caption{Strict reference-cell leaderboard. Macro-F1 is averaged over classification targets; raw RMSE over regression targets on their original scales. Elo combines within-target comparisons. Memory and time-limit failures are scored with the constant predictor; see Appendix~\ref{app:failure-rates}. Both TabPFN-Wide configurations are classification-only and cover 30 targets, not the 12 regression targets for which RMSE is measured.}
  \label{tab:leaderboard}
\end{table}

\subsection{Model order depends on the operating point and modality}

Figure~\ref{fig:sample-budget} compares six sample budgets at 10,000 features. From 20 to 1,000 training rows, the linear baseline falls from shared first place to 16th.
The lower panel recomputes Elo on the same 14 datasets across all training-sample settings. The main ranking trends persist with dataset composition held fixed: the linear baseline falls from fourth to 16th, indicating that its decline is not explained solely by changes in dataset composition.
Appendix~\ref{app:sample-composition} shows how the dataset composition changes across these settings.

Feature width also changes model order. In a separate comparison on \ContrastTargetCount{} shared targets at 100 training rows, increasing the feature budget from 2,000 to 10,000 moves TabICLv2 from rank \TabICLFeatureRankNarrow{} to \TabICLFeatureRankWide{} and Logistic Regression from \LogisticRegressionFeatureRankNarrow{} to \LogisticRegressionFeatureRankWide{}. TabICLv2 and RealTabPFN 2.5 share a regular feature limit of \TabICLFeatureLimit{}, yet RealTabPFN 2.5 remains first. Declared capacity alone does not predict the relative performance of these pipelines.

At the reference cell, RealTabPFN 2.5 leads gene expression, metagenomics, and molecular properties. TabFM leads the four embedding targets, and \ModalityLeaderMethylation{} leads the methylation targets. For genomic prediction, \ModalityLeaderGenomicPrediction{} and \ModalityRunnerUpGenomicPrediction{} lead the \ModalityTargetsGenomicPrediction{} regression targets. These pools contain at most \ModalityTargetsMetagenomics{} targets, so comparisons are descriptive. Appendix~\ref{app:modality-ranks} shows the full modality-specific rankings.

Appendix~\ref{app:fit-time} compares predictive performance and fit time.

\newpage
\section{Discussion}

TFMs generally occupy the leading ranks, but the results do not support a single default model. Although the paired bootstrap separates RealTabPFN 2.5 from TabPFN 3 at the reference cell, the leading configuration varies across biomedical modalities. The strength of Logistic Regression here differs from the pattern reported by TabArena, where foundation models excel on smaller datasets and tuning and ensembling alter model-family comparisons. These two rankings are not directly comparable: TabArena covers a broader collection of general-purpose tables and includes tuned and ensembled variants, whereas we evaluate fixed configurations on biomedical HDLSS tasks. While the ranking differs from general-purpose benchmarks, the effects of domain, metric, tuning, and ensembling cannot be fully disentangled. Scoring also plays a role: under AUROC, the metric TabArena uses for binary classification, Logistic Regression drops out of the leading group and the top \AUROCTFMLeadCount{} places are held by tabular foundation models (Appendix~\ref{app:metric-sensitivity}), so part of the contrast reflects threshold placement rather than the domain alone. One possible explanation is that pretraining provides a useful prior in small labelled cohorts, whereas regularisation keeps a linear decision rule competitive when the representation is already informative. The benchmark does not distinguish between these mechanisms.

The modality-specific comparisons show that a model should be matched to the representation and endpoint rather than taken from one overall ranking. The regression and embedding task pools remain smaller than the classification and omics task pools, so they warrant more caution.

\section{Limitations}

First, the overall ranking reflects the composition of this release, including its classification-regression balance and the uneven representation of biomedical modalities. RF shortlisting may also favour tasks learnable by forests. Manual additions broadened the collection, which spans a wide range of observed performance (Appendix~\ref{app:selection-difficulty}). The interactive leaderboard allows users to restrict comparisons to a modality of interest and inspect the corresponding performance and resource measurements. 

Second, uniform feature capping is label-agnostic and reproducible but may omit sparse biological signals. The evaluation covers five seeded feature orderings rather than the full distribution; additional orderings would multiply the already large model-dataset-budget grid. Feature-width comparisons do not estimate the value of supervised feature selection.

Third, the comparison measures default behaviour rather than performance after equal hyperparameter tuning. AutoGluon's one-hour Extreme preset was evaluated only at the reference and full-feature, full-sample cells, as extending it across the grid would substantially increase compute. Neither comparison provides an upper bound for a model family. Sample-budget comparisons may also reflect changes in class allocation and training-dependent feature filtering, rather than sample count alone.

Fourth, target coverage decreases at larger fixed sample budgets because fewer datasets contain enough training observations. The full-sample condition instead uses all available training observations for each dataset and therefore restores coverage of smaller datasets. Its target pool differs from those of the fixed-budget conditions, so ranking differences cannot be attributed to training-set size alone. The paired rating difference is likewise model-based: Bradley-Terry assigns each model a single transitive ability and does not represent the modality- and cell-dependent ordering reported above, so its interval is conditional on that assumption. Additionally, rare-class filtering narrows the target definition and precludes conclusions about underrepresented classes.  

Finally, three of the \BenchmarkDatasetCount{} benchmark datasets (OVA\_Uterus, AP\_Colon\_Kidney, and AP\_Ovary\_Lung) originate from the GEMLeR/expO collection, which also supplied OVA\_Lung and OVA\_Endometrium used in TabDPT pretraining. TabDPT performance on these related datasets should therefore be interpreted with this overlap in mind. However, neither of the RealTabPFN models has any overlap.
\newpage
\section{Conclusion}

By evaluating tabular learners across heterogeneous biomedical measurements with many features and few samples, TabBench-Bio exposes behaviour that general-purpose suites rarely measure. Across the \BenchmarkDatasetCount-dataset benchmark, TFMs generally occupy the leading ranks, and RealTabPFN 2.5 leads the reference-cell point ranking, followed by TabPFN 3 and Logistic Regression, whose point estimates are nearly identical. A paired bootstrap separates RealTabPFN 2.5 from TabPFN 3 at the reference cell, while modality-specific results show that the strongest configuration depends on the biological representation. Researchers should consider performance under conditions relevant to their data when selecting a model. TabBench-Bio supports this comparison and helps developers identify where their models perform reliably.

\section{Outlook and Living Benchmark}
\label{sec:living-benchmark}

TabBench-Bio is designed as a living benchmark. Since its initial release, we have added tuned classical baselines and models not covered by the main comparison in this paper, and we continue to add models and datasets. We welcome submissions of biomedical tabular datasets, particularly from underrepresented assays and clinical endpoints, for future releases.

The results presented here cover the submitted collection of \BenchmarkDatasetCount{} datasets and \PredictivePeerModelCount{} model configurations; subsequent additions are tracked separately on the live leaderboard. This separation maintains a fixed reference point for the paper while letting the benchmark itself evolve.

\begin{figure}[t]
  \centering
  \includegraphics[width=\linewidth]{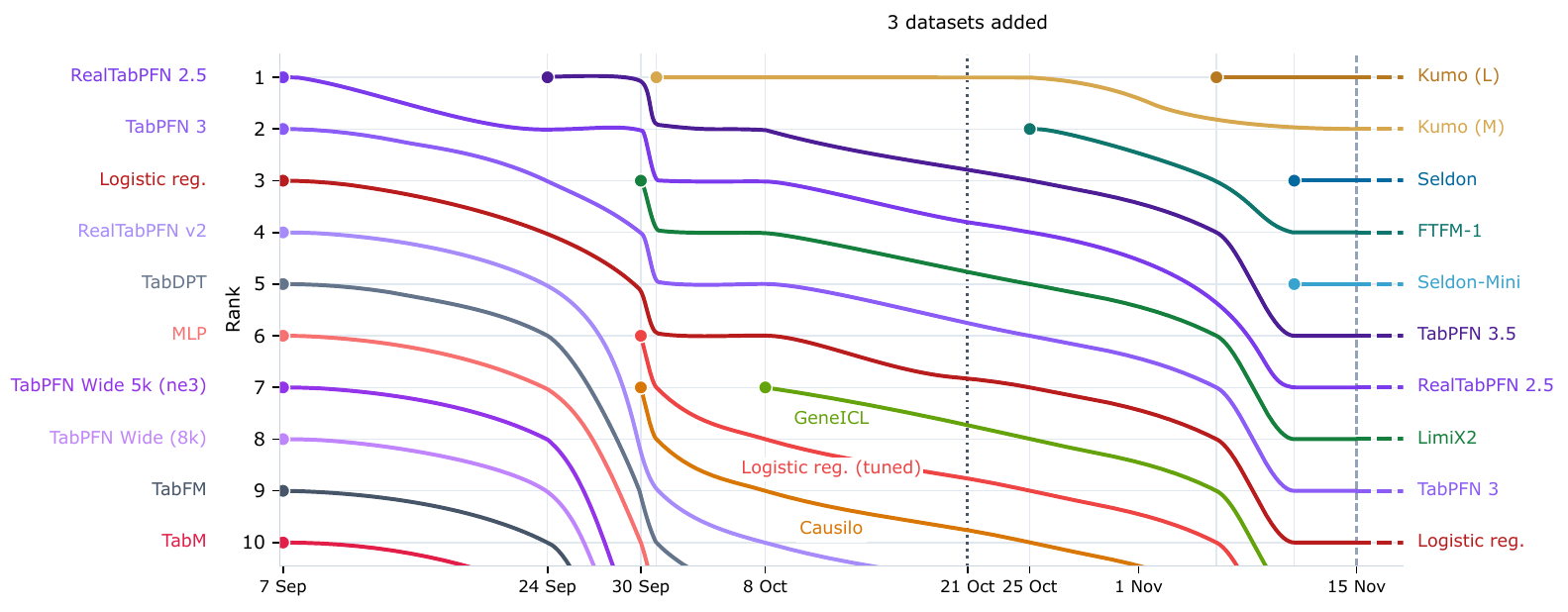}
  \caption{Evolution of the top ten as models are added at the strict reference cell (10,000-feature cap, 100 training samples). Dates indicate when models were added to the benchmark, not when they were released.}
  \label{fig:living-benchmark}
\end{figure}

Figure~\ref{fig:living-benchmark} shows how successive model releases reshuffle the leading group at the strict reference cell. Several newer foundation models rank above the strongest model in the submitted comparison, and foundation models continue to dominate the top positions. Because the leading model changes between releases, sustained evaluation under a common protocol is essential. The timeline reconstructs rankings from current results and model availability rather than measuring performance retroactively at each historical date.
\newpage
\section*{Reproducibility and Data availability}

The benchmark code, versioned registry, reporting scripts, and released result snapshots are available at \url{https://github.com/not-a-feature/TabBench-Bio}. The interactive leaderboard and machine-readable exports are served at \url{https://tabbench-bio.eu}.
The locally generated embedding datasets are publicly available, including target labels, cross-validation group identifiers, and scripts documenting their construction.
For each release, an aggregation manifest with cryptographic digests links the raw predictions and terminal status records to the generated tables, figures, and manuscript values.
The evaluation protocol is described in Section~\ref{sec:benchmark}, with further details on model defaults, failure rates, the dataset registry, and local embedding construction in Appendices~\ref{app:model-defaults}--\ref{app:local-construction}. Splits are frozen and reused across all models and operating points. Aggregation is deterministic, and the manifest described above links all recorded predictions and status records to the results in this manuscript.

\section*{Ethics statement}

This work evaluates published models on existing secondary datasets, including locally generated sequence-embedding datasets derived from existing biological sequences. No new human-subject data were collected and no institutional review board approval was required. No attempt at re-identification is made. Several sources derive from human cohorts, so the results describe default model behaviour on retrospective research data. The benchmark does not establish clinical validity, and no ranking reported here supports use in patient care.

\section*{AI use statement}

In this work, we used generative AI tools, specifically LLMs (mainly Claude Opus 5 and GPT-6 Astra) for coding and writing assistance. 
We have not used generative AI tools to propose or refine the research questions and hypotheses, to design the benchmark protocol, the operating-point grid, or the aggregation and rating methodology, to generate synthetic data, to formulate or prove mathematical claims, or to interpret the reported results; these were all carried out by the authors.

They were used to assist in implementing code by providing suggestions and generating code snippets, but not for full code generation.
We used them for multiple code review steps. Additionally, we have reviewed all AI-assisted work. AI-assisted code was read by at least two authors and is covered by the repository's automated test suite. We take responsibility for the final content of this work, including all text, claims, and artefacts produced.

\section*{Author contributions}

JK conceived the study, developed the benchmark. JB developed the automated dataset preselection and was supervised by JK and JH. JK, JH and JB contributed to dataset curation and selection. NP supervised the project. JK, JH, SO and NP contributed to writing.

\bibliographystyle{iclr2027_conference}
\bibliography{tabbench_bio}

\appendix
\setcounter{figure}{0}
\renewcommand{\thefigure}{A\arabic{figure}}
\renewcommand{\theHfigure}{A\arabic{figure}}
%
%

\newpage
\section{Model configuration defaults}
\label{app:model-defaults}

Table~\ref{tab:model-defaults} lists the key defaults of the benchmark model configurations. Maximum iteration counts are subject to model-native early stopping and the common one-hour timeout. Peer models use no hyperparameter optimisation or AutoGluon bagging; AutoGluon Extreme is an external reference. Parameters not listed remain at the corresponding library defaults.

\begingroup
\footnotesize
\setlength{\tabcolsep}{5pt}
\renewcommand{\arraystretch}{0.96}
\begin{longtable}{@{}p{0.24\linewidth}p{0.70\linewidth}@{}}
\toprule
Model & Important default parameters \\
\midrule
\endfirsthead
\toprule
Model & Important default parameters \\
\midrule
\endhead
\midrule
\multicolumn{2}{r}{\footnotesize Continued on next page} \\
\endfoot
\bottomrule
\addlinespace[6pt]
\caption{Principal default parameters of the benchmark model configurations. Parameters not listed remain at the corresponding library defaults.}
\label{tab:model-defaults} \\
\endlastfoot
Constant & Class-prior predictor (classification); mean predictor (regression) \\
CatBoost & Learning rate $0.05$; at most $10{,}000$ iterations; adaptive early stopping \\
LightGBM & Learning rate $0.05$; at most $10{,}000$ rounds; adaptive early stopping \\
KNN & $k=5$; uniform weights; Euclidean distance \\
Logistic Regression & $L_2$ penalty; $C=1$ (classification); $\alpha=1$ (regression); standardised inputs \\
Random Forest & $300$ trees; bootstrap sampling; at most $15{,}000$ leaf nodes \\
XGBoost & Tree booster; learning rate $0.1$; at most $10{,}000$ rounds; adaptive early stopping \\
Extra Trees & $300$ trees; bootstrap sampling; at most $15{,}000$ leaf nodes \\
MLP & Four layers of width $128$; ReLU; dropout $0.1$; Adam; learning rate $3\times10^{-4}$; at most $1{,}000$ epochs \\
RealMLP & TD preset; three layers of width $256$; $256$ epochs; learning rate $0.04$ (classification) or $0.2$ (regression) \\
TabM & TabM-mini; $32$ subnetworks; block width $512$; dropout $0.1$; learning rate $2\times10^{-3}$; patience $16$ \\
MITRA & Tab2D; one estimator; $50$ fine-tuning steps; learning rate $10^{-4}$; patience $40$ \\
RealTabPFN v2.0 & Default v2 checkpoint; $8$ estimators; $2$ features per group; temperature $0.9$ \\
RealTabPFN 2.5 & Default v2.5 checkpoint; $8$ estimators; $3$ features per group; temperature $0.9$ \\
TabPFN 3 & Default v3 checkpoint; $8$ estimators; feature-group size $3$; temperature $0.9$ \\
TabPFN Wide (8k) & \texttt{wide-v2-8k} checkpoint; one estimator; one feature per group \\
TabPFN Wide 5k (ne3) & \texttt{wide-v2-5k} checkpoint; $3$ estimators; one feature per group \\
TabFM & Default v1.0 PyTorch checkpoint; $32$ estimators; random feature permutations; no feature cap; averaged logits \\
TabDPT & Default v1.2 checkpoint; $8$ ensembles; full training context; standard normalisation; PCA feature reduction; temperature $1$ \\
TabICLv2 & TabICLv2 checkpoint; $8$ estimators; Latin feature shuffling; shifted classes; averaged logits \\
AutoGluon & Extreme preset; one-hour budget; native model portfolio and weighted ensemble \\
\end{longtable}

\endgroup

\newpage
\section{Performance and fit time}
\label{app:fit-time}

At the reference cell, RealTabPFN 2.5 has a median fit time of \TopModelMedianFitSeconds{} seconds per fold. \FastCompetitiveModel{} ranks first in mean macro-F1 (third in Elo) with a median fit time of just \FastCompetitiveFitSeconds{} seconds. Fit time measures the wall-clock duration of the adapter's fit call: training or fine-tuning for models that update their parameters, and preparation of the training context for in-context learning models. Times cover successful fits and include the shared harness overhead. They exclude prediction, pretraining, memory, and energy costs. The dashed frontier joins methods for which no other plotted method is both at least as accurate and faster to fit.

\begin{figure}[htbp]
  \centering
  \includegraphics[width=0.7\linewidth]{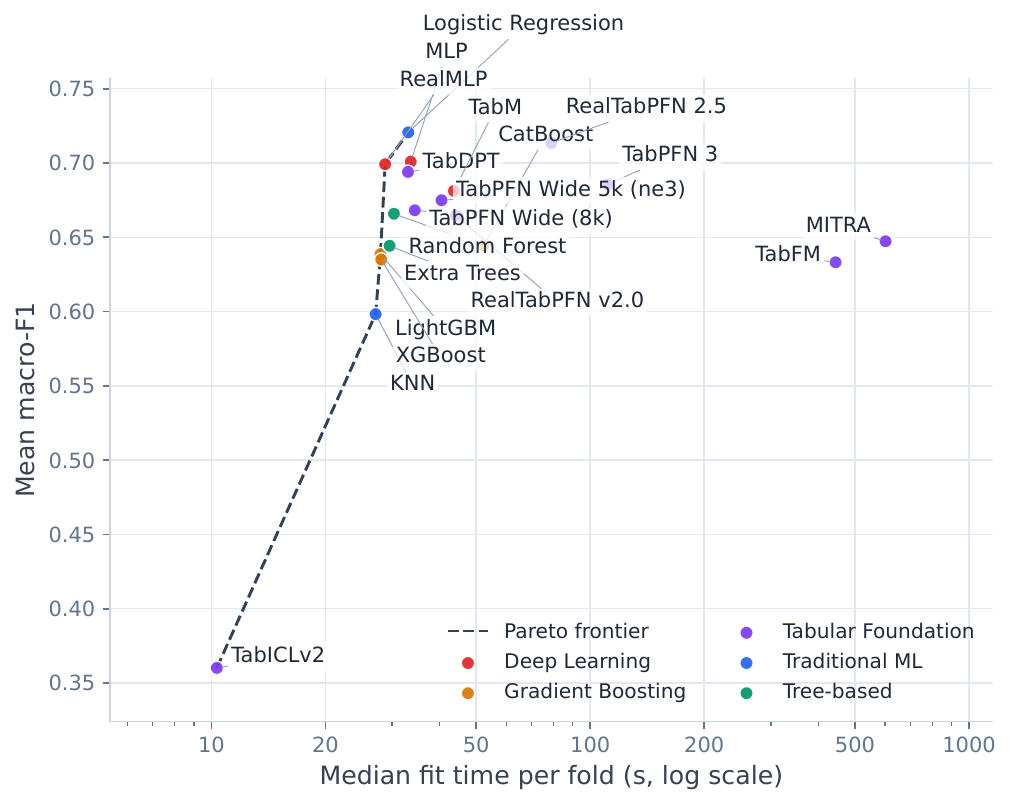}
  \caption{Mean macro-F1 against median fit time at \ReferenceCell{}. The time axis is logarithmic.}
  \label{fig:cost}
\end{figure}
\FloatBarrier

\newpage
\section{Full-feature, full-sample ranking}
\label{app:full-cell}

\begin{figure}[!ht]
  \centering
  \includegraphics[width=0.94\linewidth]{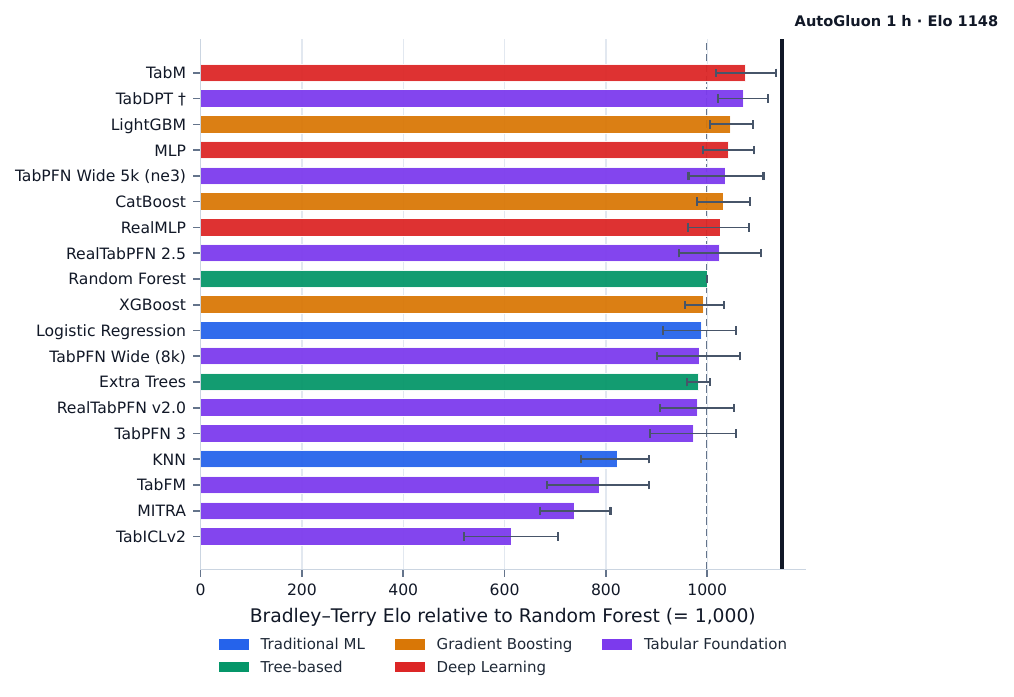}
  \caption{Strict fold-level Elo using all available training rows and features after shared preprocessing. Most models cover 43 targets; the classification-only TabPFN-Wide models cover 30. Error bars show 95\% target-bootstrap intervals relative to Random Forest (= 1,000). The solid vertical line marks the one-hour AutoGluon Extreme reference. \dag{}~TabDPT: Related datasets derived from the same collection were included in TabDPT pretraining.}
  \label{fig:elo-full}
\end{figure}
\FloatBarrier

\newpage
\section{Leading models across the grid}
\label{app:grid-winners}

\begin{figure}[!ht]
  \centering
  \includegraphics[width=0.98\linewidth]{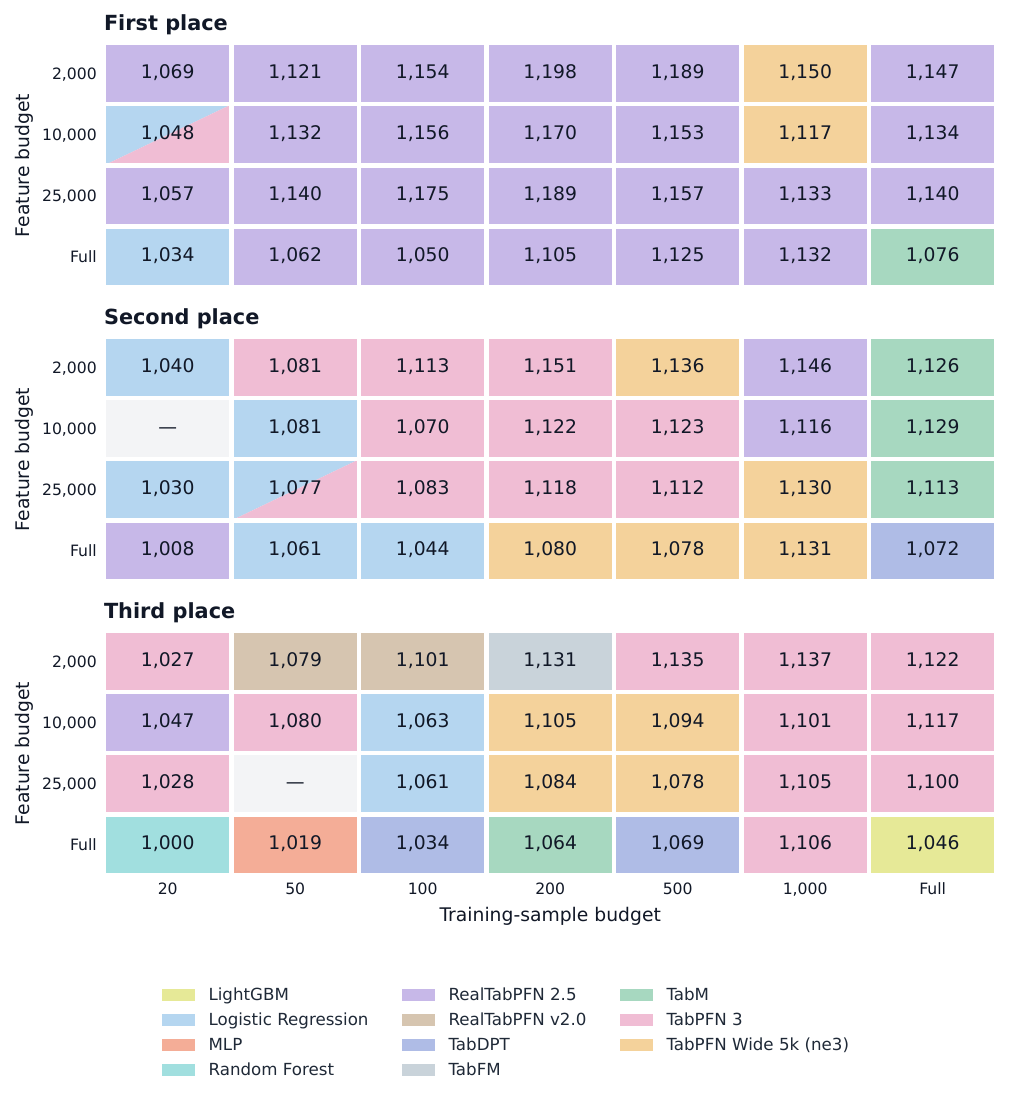}
  \caption{First, second and third place at each grid point under strict fold-level Elo. Colours identify models; numbers give rounded bootstrap-median Elo. Split cells denote ties; subsequent ranks are skipped and marked by a dash. AutoGluon is excluded as an external reference. Target coverage varies across cells; rank differences do not establish statistical significance. Full uses all available training rows or features after shared preprocessing.}
  \label{fig:grid-winners}
\end{figure}
\FloatBarrier

\clearpage
\section{Modality-specific model rankings}
\label{app:modality-ranks}

\begin{figure}[!ht]
  \centering
  \includegraphics[width=\linewidth]{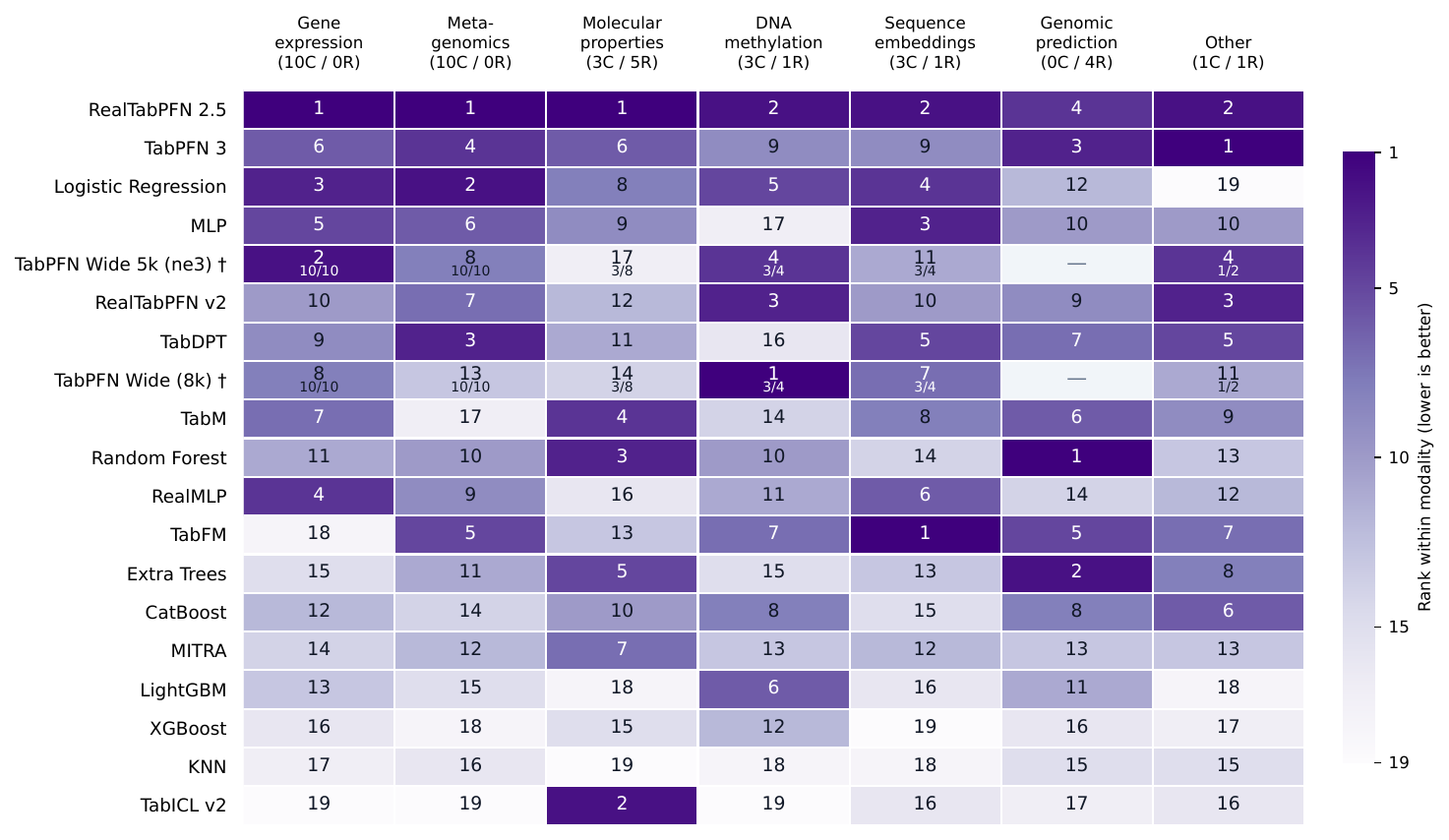}
  \caption{Within-modality ranks at the strict reference cell (10,000-feature cap, 100 training samples), using median Elo from 2,000 target bootstraps. C/R denote classification/regression dataset counts. \dag{} marks classification-only models; fractions give covered/total datasets, and dashes indicate no applicable tasks. The constant baseline and AutoGluon are excluded from displayed ranks.}
  \label{fig:modality-ranks}
\end{figure}

Figure~\ref{fig:modality-ranks} compares model order within the original modality groups. The small pools and differences in task type and model coverage make these rankings descriptive rather than evidence of a universal ordering for new datasets.

\FloatBarrier

\newpage
\section{Dataset coverage and retained feature dimensions}
\label{app:sample-composition}

\begin{figure}[!ht]
  \centering
  \includegraphics[width=\linewidth]{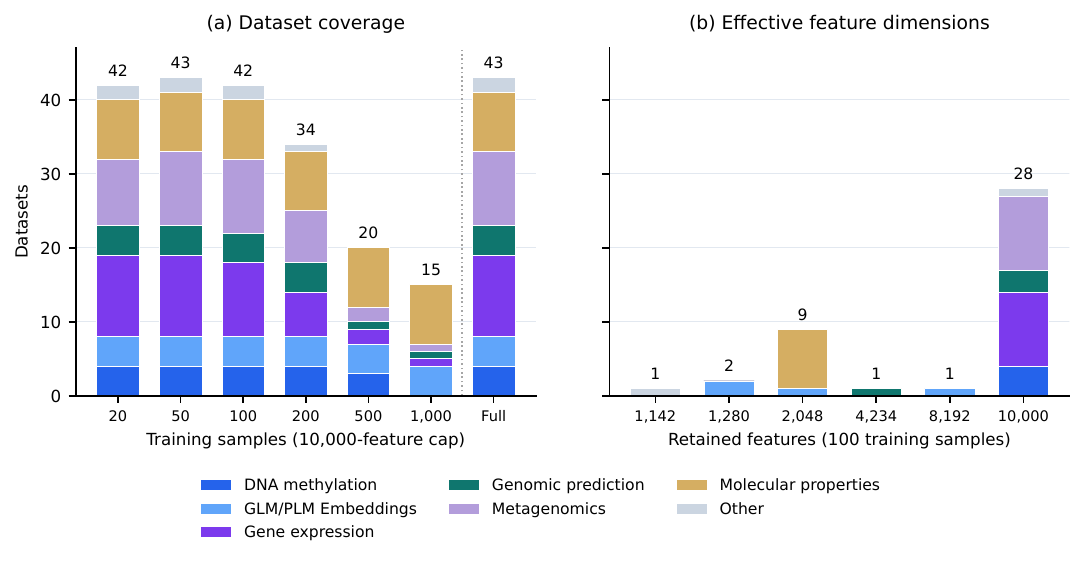}
  \caption{Dataset counts by modality. (a) Coverage in the strict results across training-sample settings at a 10,000-feature cap; failure-imputed results are retained. Full uses each fold's available training rows. (b) Distribution of the 42 reference-cell datasets by the number of features retained after shared training-data filtering and the 10,000-feature cap, using 100 training samples. Retained counts agree across all five folds; 28 datasets reach the cap. Each dataset is counted once per sample setting in (a) and once in (b); numbers above bars give dataset totals.}
  \label{fig:sample-composition}
\end{figure}

Figure~\ref{fig:sample-composition} shows dataset coverage across sample settings and retained feature counts at the reference cell. Larger fixed sample budgets exclude smaller datasets as not enough samples are available; Full uses all available training rows within each fold. The feature cap is an upper limit: datasets with fewer available features remain narrower.

The matched pool in Figure~\ref{fig:sample-budget} contains the 14 datasets available at every fixed training-sample setting: seven classification and seven regression tasks, comprising eight molecular-property, four sequence-embedding, one gene-expression and one genomic-prediction dataset. It excludes methylation and metagenomics, so its trends describe a narrower task pool. Model applicability restrictions are unchanged.

\FloatBarrier

\newpage
\section{Metric sensitivity}
\label{app:metric-sensitivity}

\begin{figure}[!ht]
  \centering
  \includegraphics[width=0.94\linewidth]{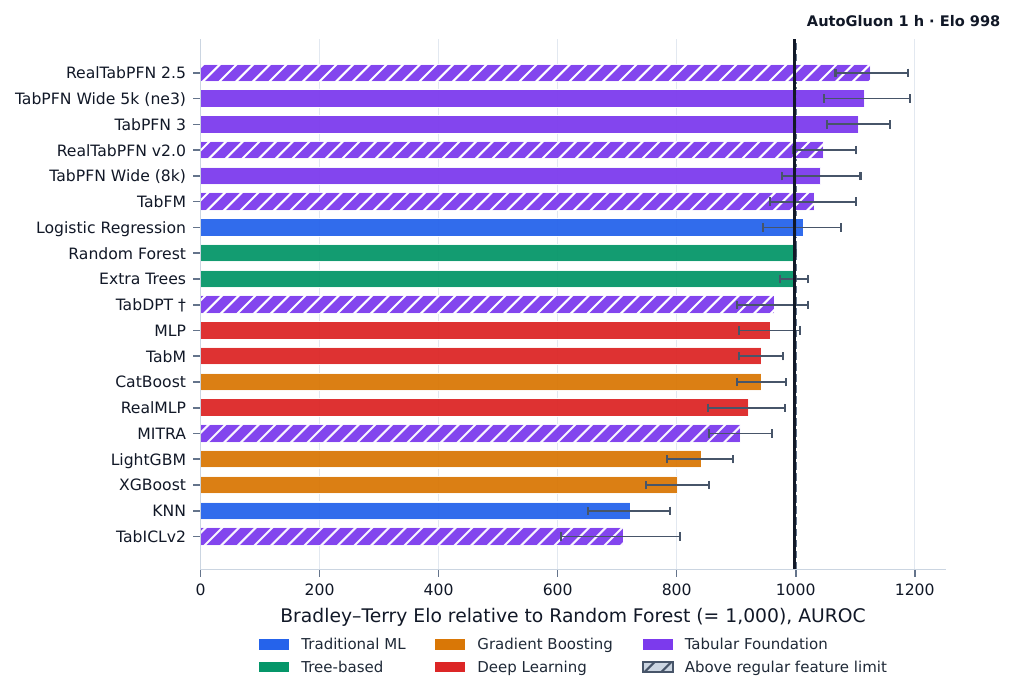}
  \caption{Reference-cell ranking ordered by AUROC. Identical runs, folds, and estimator as Figure~\ref{fig:elo}; only the classification score changes, recomputed from the stored fold-level predictions without refitting. AUROC is applied to every classification target, one-vs-rest for the multiclass ones. This is a post-hoc metric sensitivity analysis using fits selected or early-stopped based on macro-F1; AUROC-based selection could change the results, and unchanged predictions do not imply unchanged rankings. AutoGluon Extreme additionally selects its ensemble weights on macro-F1 and is reported as a reference rather than a ranked peer.}
  \label{fig:elo-auroc}
\end{figure}

Macro-F1 combines discrimination with threshold placement; decision-threshold handling follows the AutoGluon adapter. Figure~\ref{fig:elo-auroc} re-orders the reference cell by AUROC as a sensitivity check, keeping regression on RMSE. AUROC is also the metric TabArena uses for binary classification, so the figure shows how much of the contrast with that benchmark depends on scoring.

The ranking changes. \AUROCTopModel{} has the highest AUROC point estimate, ahead of \AUROCSecondModel{} and \AUROCThirdModel{}, with the three separated by \AUROCLeadingSpread{} Elo. The larger change is below them: Logistic Regression, third on macro-F1, falls behind the leading \AUROCTFMLeadCount{} places, all of which are held by tabular foundation models. Comparisons between model families must therefore be read together with the metric under which they are
computed.

\FloatBarrier

\newpage
\section{Recorded failure rates}
\label{app:failure-rates}

Table~\ref{tab:failure-reference} reports the recorded terminal outcomes per model at the reference cell. Memory and time-limit failures are scored with the constant predictor and therefore enter the rankings in Figure~\ref{fig:elo} and Table~\ref{tab:leaderboard}; design skips do not. Failure counts are concentrated in a few models rather than spread across the roster, so the deployability component of the ranking affects some models far more than others.

\begin{table}[!htbp]
\centering
\begingroup
\footnotesize
\setlength{\tabcolsep}{3pt}
\begin{tabular}{lrrrrrr}
\toprule
Model & Pass/attempt & Fail (\%) & Memory & Timeout & Other & Skip \\
\midrule
Constant & 210/210 & 0.0 & 0 & 0 & 0 & 5 \\
CatBoost & 203/210 & 3.3 & 0 & 7 & 0 & 5 \\
LightGBM & 210/210 & 0.0 & 0 & 0 & 0 & 5 \\
KNN & 160/210 & 23.8 & 0 & 0 & 50 & 5 \\
Logistic Regression & 210/210 & 0.0 & 0 & 0 & 0 & 5 \\
Random Forest & 210/210 & 0.0 & 0 & 0 & 0 & 5 \\
XGBoost & 210/210 & 0.0 & 0 & 0 & 0 & 5 \\
Extra Trees & 210/210 & 0.0 & 0 & 0 & 0 & 5 \\
MLP & 210/210 & 0.0 & 0 & 0 & 0 & 5 \\
RealMLP & 210/210 & 0.0 & 0 & 0 & 0 & 5 \\
TabM & 210/210 & 0.0 & 0 & 0 & 0 & 5 \\
MITRA & 205/210 & 2.4 & 0 & 0 & 5 & 5 \\
RealTabPFN v2.0 & 205/210 & 2.4 & 0 & 0 & 5 & 5 \\
RealTabPFN 2.5 & 205/210 & 2.4 & 0 & 0 & 5 & 5 \\
TabPFN 3 & 205/210 & 2.4 & 0 & 0 & 5 & 5 \\
TabPFN Wide (8k) & 145/150 & 3.3 & 0 & 0 & 5 & 65 \\
TabPFN Wide 5k (ne3) & 145/150 & 3.3 & 0 & 0 & 5 & 65 \\
TabFM & 205/210 & 2.4 & 0 & 0 & 5 & 5 \\
TabDPT & 210/210 & 0.0 & 0 & 0 & 0 & 5 \\
TabICLv2 & 55/210 & 73.8 & 155 & 0 & 0 & 5 \\
AutoGluon (ref.) & 210/210 & 0.0 & 0 & 0 & 0 & 5 \\
\bottomrule
\end{tabular}
\endgroup
\caption{Recorded failure rates at p=10,000, n=100. Counts are recorded terminal dataset--fold outcomes at the nominal cell, before adaptive fallback. Failure percentage is $100F/(P+F)$; design skips are excluded from its denominator. Memory denotes recorded \texttt{fit\_oom} or \texttt{inference\_oom}, including possible estimator refusals; timeout denotes \texttt{time\_limit}. Other includes generic errors and unspecified reasons; recorded codes do not establish root causes. Category and skip columns are counts. A dash denotes no attempts or no recorded outcomes, not zero failure. Missing runs are not counted; this table does not certify completion. Models may have different eligible task pools. AutoGluon is a separate reference. }
\label{tab:failure-reference}
\end{table}

\FloatBarrier
\newpage
\subsection{Failure-policy sensitivity}
\label{app:failure-policy}

\begin{figure}[!ht]
  \centering
  \includegraphics[width=0.94\linewidth]{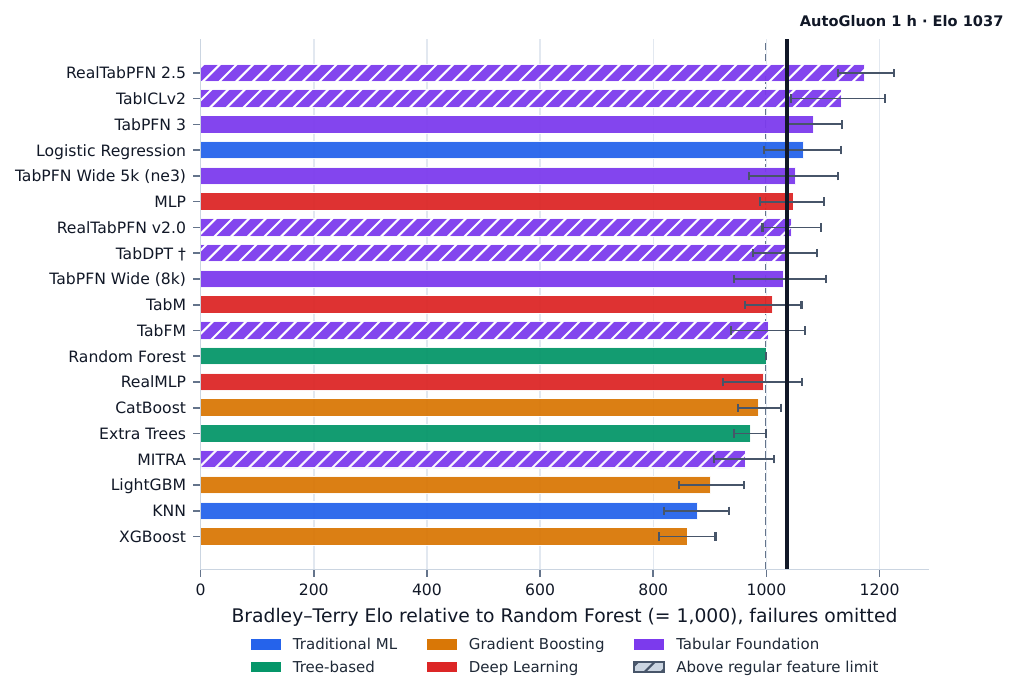}
  \caption{Reference-cell ranking with failures omitted. Runs, folds, metric and estimator are unchanged from Figure~\ref{fig:elo}; no models are refitted. Omitting failures removes affected model--target pairs, so target coverage differs between models. \dag{}~TabDPT: Related datasets derived from the same collection were included in TabDPT pretraining.}
  \label{fig:elo-conditional}
\end{figure}

The primary ranking scores failures at the constant predictor. Figure~\ref{fig:elo-conditional} instead omits failures and retains only model--target pairs with all scheduled folds scored. Models therefore cover different target pools. This sensitivity analysis does not establish bounds on ratings or ranks.

The rank correlation between the two policies is \PolicyRankCorrelation{}, and \PolicyMovedModelCount{} models change rank. \PolicyLargestMoveModel{} moves the most: \PolicyLargestMoveRanks{} places after losing \PolicyLargestMoveLostTargets{} targets. \PolicyTopModel{} still leads, followed by \PolicySecondModel{}.

\FloatBarrier

\newpage
\section{Dataset screening and observed difficulty}
\label{app:selection-difficulty}

\begin{figure}[!ht]
  \centering
  \includegraphics[width=\linewidth]{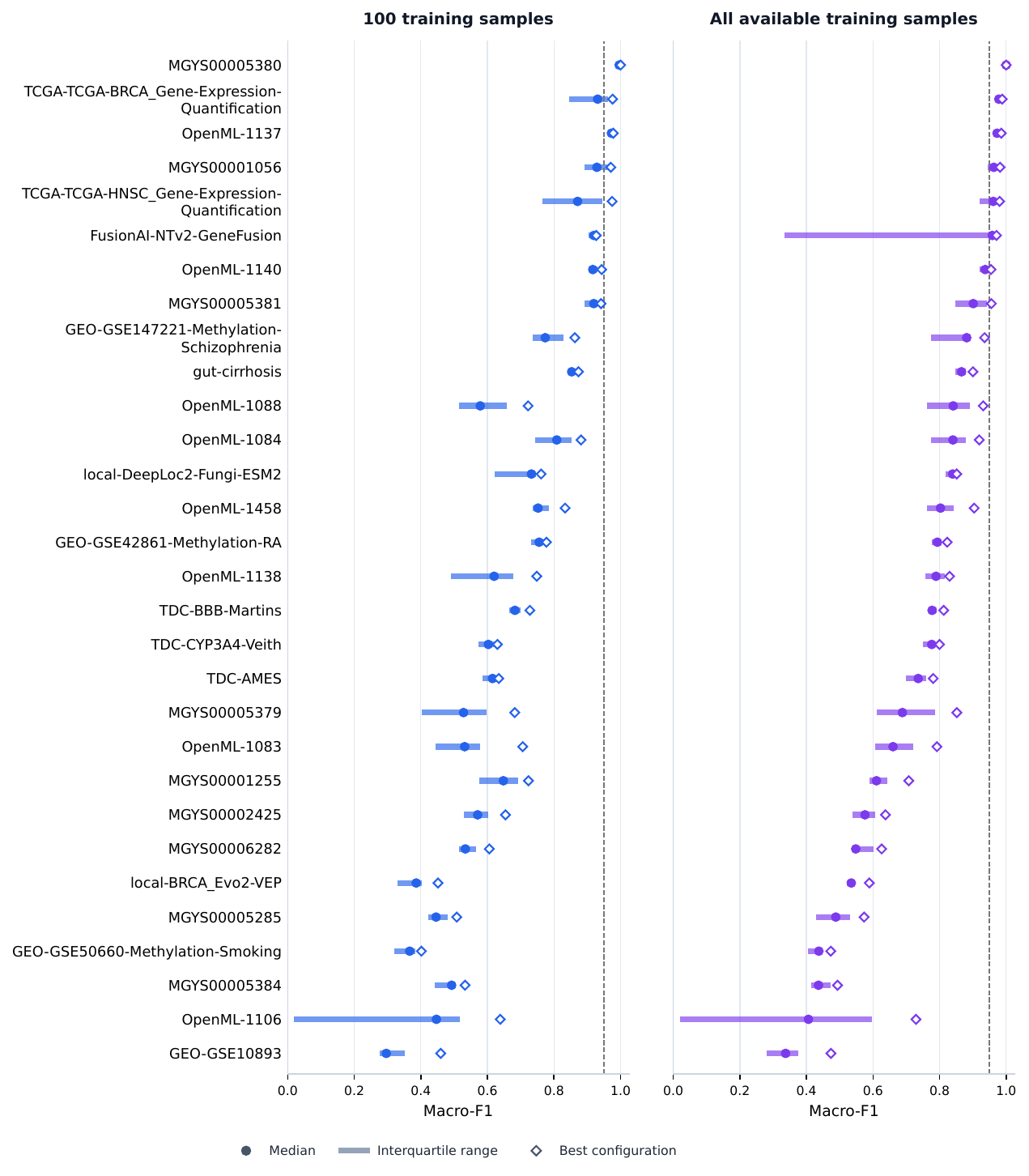}
  \caption{Classification performance on the 30 datasets at a 10,000-feature cap, with 100 training samples (left) or all available training samples (right). Circles, bars and diamonds show the median, interquartile range and best five-fold mean macro-F1 across the 19 model configurations. Bars are not confidence intervals. Full sample uses each dataset's available training rows. Strict failure-imputed scores are included; DUMMY and AutoGluon are excluded. The dashed line marks \(0.95\).}
  \label{fig:classification-difficulty}
\end{figure}

\begin{figure}[!ht]
  \centering
  \includegraphics[width=\linewidth]{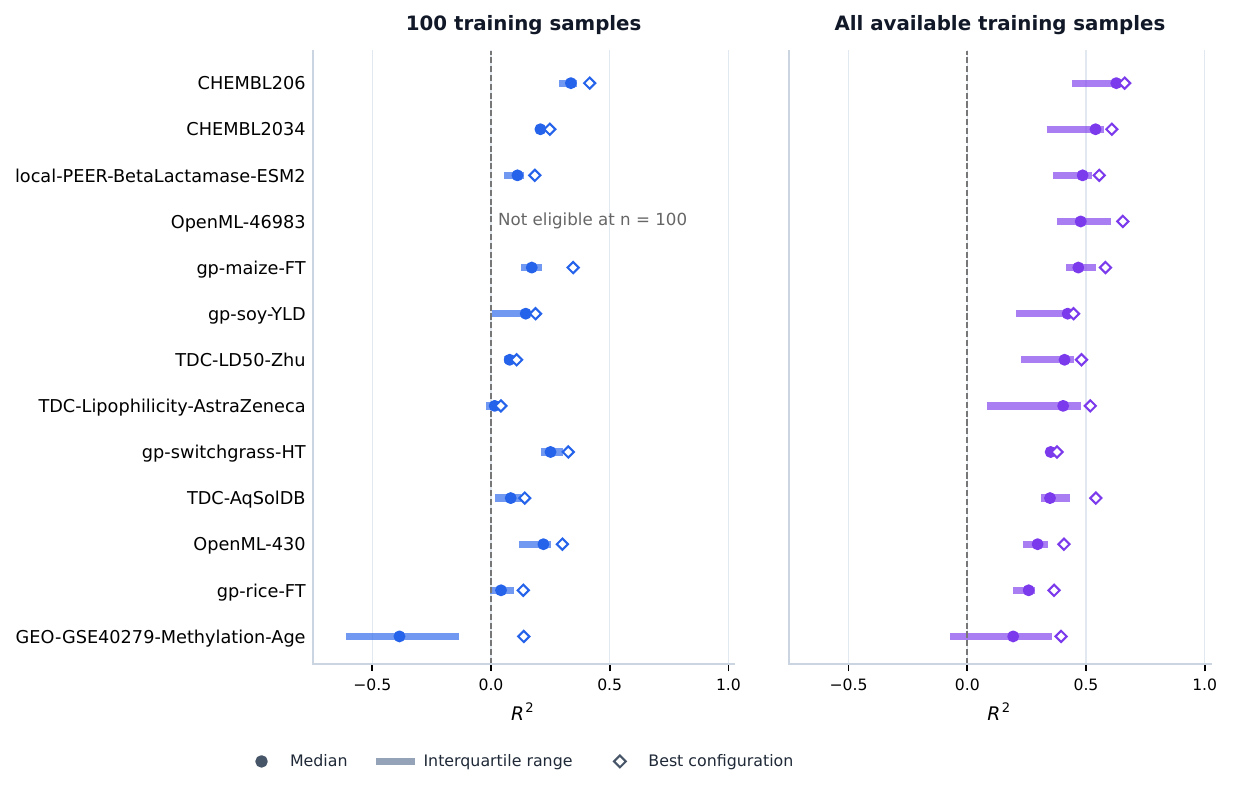}
  \caption{Regression performance at a 10,000-feature cap, with 100 training samples (left) or all available training samples (right). Circles, bars and diamonds show the median, interquartile range and best five-fold mean \(R^2\) across the 17 model configurations. \nolinkurl{OpenML-46983} has too few training rows for the 100-sample setting. Full sample uses each dataset's available training rows. Strict failure-imputed scores are included; DUMMY and AutoGluon are excluded. The dashed line marks \(R^2=0\).}
  \label{fig:regression-difficulty}
\end{figure}

Figures~\ref{fig:classification-difficulty} and \ref{fig:regression-difficulty} summarize per-dataset performance under the strict failure policy.
At 100 training samples, median macro-F1 reaches \(0.95\) on two of 30 classification tasks, and the best configuration reaches it on five. With all available training rows, these counts rise to six and eight. Only one dataset, \nolinkurl{MGYS00005380}, exceeds macro-F1 \(0.99\) in either setting, for both the median and best configuration. No regression task reaches best mean \(R^2\geq0.95\); the maximum is \(0.416\) at 100 samples and \(0.663\) at full sample.  A few tasks approach saturation, particularly with more training data, but the benchmark as a whole spans a broad range of predictive performance.

\FloatBarrier
\newpage

\FloatBarrier

\newpage
\section{Dataset registry}
\label{app:datasets}

Table~\ref{tab:datasets} is generated from the dataset configuration and versioned registry used for the benchmark. Registry IDs remain stable even when a source title is abbreviated; the generated metadata CSV also retains source and licence fields. Samples counts the rows the benchmark evaluates, after classes with fewer than ten samples are removed. Features is the native width of the source table, before any feature cap is applied, so a dataset wider than a cell's budget contributes only a random subset of these columns in that cell.
\begingroup
\fontsize{8}{8.5}\selectfont
\setlength{\tabcolsep}{1.8pt}
\renewcommand{\arraystretch}{0.92}
\begin{longtable}{@{}>{\raggedright\arraybackslash}p{0.24\linewidth}>{\raggedright\arraybackslash}p{0.365\linewidth}>{\raggedright\arraybackslash}p{0.095\linewidth}>{\raggedright\arraybackslash}p{0.045\linewidth}rrc@{}}
\toprule
Registry ID & Dataset & Modality & Task & Samples & Features & \\
\midrule
\endfirsthead
\toprule
Registry ID & Dataset & Modality & Task & Samples & Features & \\
\midrule
\endhead
\bottomrule
\addlinespace[6pt]
\caption{Dataset registry. C: classification; R: regression. Arrows link to dataset sources; for locally constructed embeddings, they link to Appendix~\ref{app:local-construction}.}
\label{tab:datasets} \\
\endlastfoot
\nolinkurl{OpenML-1138} & Uterus vs. Other Tissue & Expr. & C & 1,545 & 10,935 & \href{https://www.openml.org/search?type=data&sort=runs&id=1138&status=active}{\(\scriptstyle\nearrow\)} \\
\nolinkurl{OpenML-1458} & ARCENE Serum Cancer & Other & C & 200 & 10,000 & \href{https://www.openml.org/search?type=data&sort=runs&id=1458&status=active}{\(\scriptstyle\nearrow\)} \\
\nolinkurl{TCGA-TCGA-BRCA_Gene-Expression-Quantification} & TCGA Breast (BRCA, RNA-seq) & Expr. & C & 1,231 & 60,660 & \href{https://portal.gdc.cancer.gov/projects/TCGA-BRCA}{\(\scriptstyle\nearrow\)} \\
\nolinkurl{GEO-GSE10893} & Breast Cancer Subtypes (GSE10893) & Expr. & C & 165 & 109,315 & \href{https://www.ncbi.nlm.nih.gov/geo/query/acc.cgi?acc=GSE10893}{\(\scriptstyle\nearrow\)} \\
\nolinkurl{TCGA-TCGA-HNSC_Gene-Expression-Quantification} & TCGA Head \& Neck (HNSC, RNA-seq) & Expr. & C & 566 & 60,660 & \href{https://portal.gdc.cancer.gov/projects/TCGA-HNSC}{\(\scriptstyle\nearrow\)} \\
\nolinkurl{local-BRCA_Evo2-VEP} & BRCA VEP (Evo2) & Embed. & C & 3,893 & 8,192 & \hyperref[app:local-construction]{\(\scriptstyle\nearrow\)} \\
\nolinkurl{local-DeepLoc2-Fungi-ESM2} & DeepLoc2 Fungi (ESM-2) & Embed. & C & 5,841 & 1,280 & \hyperref[app:local-construction]{\(\scriptstyle\nearrow\)} \\
\nolinkurl{OpenML-1083} & mouseType & Expr. & C & 214 & 45,101 & \href{https://www.openml.org/search?type=data&sort=runs&id=1083&status=active}{\(\scriptstyle\nearrow\)} \\
\nolinkurl{OpenML-1106} & GCM & Expr. & C & 190 & 16,063 & \href{https://www.openml.org/search?type=data&sort=runs&id=1106&status=active}{\(\scriptstyle\nearrow\)} \\
\nolinkurl{OpenML-1137} & AP\_Colon\_Kidney & Expr. & C & 546 & 10,935 & \href{https://www.openml.org/search?type=data&sort=runs&id=1137&status=active}{\(\scriptstyle\nearrow\)} \\
\nolinkurl{OpenML-1084} & BurkittLymphoma & Expr. & C & 220 & 22,283 & \href{https://www.openml.org/search?type=data&sort=runs&id=1084&status=active}{\(\scriptstyle\nearrow\)} \\
\nolinkurl{OpenML-1088} & variousCancers\_final & Expr. & C & 383 & 54,675 & \href{https://www.openml.org/search?type=data&sort=runs&id=1088&status=active}{\(\scriptstyle\nearrow\)} \\
\nolinkurl{OpenML-1140} & AP\_Ovary\_Lung & Expr. & C & 324 & 10,935 & \href{https://www.openml.org/search?type=data&sort=runs&id=1140&status=active}{\(\scriptstyle\nearrow\)} \\
\nolinkurl{MGYS00005384} & MGnify functional profile & Metag. & C & 674 & 17,026 & \href{https://www.ebi.ac.uk/metagenomics/studies/MGYS00005384}{\(\scriptstyle\nearrow\)} \\
\nolinkurl{MGYS00005380} & MGnify functional profile & Metag. & C & 264 & 16,690 & \href{https://www.ebi.ac.uk/metagenomics/studies/MGYS00005380}{\(\scriptstyle\nearrow\)} \\
\nolinkurl{MGYS00005285} & MGnify functional profile & Metag. & C & 438 & 12,705 & \href{https://www.ebi.ac.uk/metagenomics/studies/MGYS00005285}{\(\scriptstyle\nearrow\)} \\
\nolinkurl{MGYS00006282} & MGnify functional profile & Metag. & C & 256 & 12,968 & \href{https://www.ebi.ac.uk/metagenomics/studies/MGYS00006282}{\(\scriptstyle\nearrow\)} \\
\nolinkurl{MGYS00002425} & MGnify functional profile & Metag. & C & 394 & 12,651 & \href{https://www.ebi.ac.uk/metagenomics/studies/MGYS00002425}{\(\scriptstyle\nearrow\)} \\
\nolinkurl{MGYS00005381} & MGnify functional profile & Metag. & C & 218 & 13,394 & \href{https://www.ebi.ac.uk/metagenomics/studies/MGYS00005381}{\(\scriptstyle\nearrow\)} \\
\nolinkurl{MGYS00001255} & MGnify functional profile & Metag. & C & 152 & 14,103 & \href{https://www.ebi.ac.uk/metagenomics/studies/MGYS00001255}{\(\scriptstyle\nearrow\)} \\
\nolinkurl{MGYS00001056} & MGnify functional profile & Metag. & C & 306 & 15,268 & \href{https://www.ebi.ac.uk/metagenomics/studies/MGYS00001056}{\(\scriptstyle\nearrow\)} \\
\nolinkurl{MGYS00005379} & MGnify functional profile & Metag. & C & 1,450 & 14,826 & \href{https://www.ebi.ac.uk/metagenomics/studies/MGYS00005379}{\(\scriptstyle\nearrow\)} \\
\nolinkurl{gut-cirrhosis} & Gut microbiome markers (healthy vs cirrhosis) & Metag. & C & 232 & 33,539 & \href{https://github.com/segatalab/metaml}{\(\scriptstyle\nearrow\)} \\
\nolinkurl{TDC-BBB-Martins} & Blood-Brain Barrier Penetration (Martins) & Mol. & C & 1,954 & 2,048 & \href{https://dataverse.harvard.edu/api/access/datafile/4259566}{\(\scriptstyle\nearrow\)} \\
\nolinkurl{TDC-AMES} & Ames Mutagenicity & Mol. & C & 7,255 & 2,048 & \href{https://dataverse.harvard.edu/api/access/datafile/4259564}{\(\scriptstyle\nearrow\)} \\
\nolinkurl{TDC-CYP3A4-Veith} & CYP3A4 Inhibition (Veith) & Mol. & C & 12,300 & 2,048 & \href{https://dataverse.harvard.edu/api/access/datafile/4259582}{\(\scriptstyle\nearrow\)} \\
\nolinkurl{GEO-GSE42861-Methylation-RA} & Rheumatoid Arthritis DNA Methylation (GSE42861) & Meth. & C & 689 & 30,000 & \href{https://www.ncbi.nlm.nih.gov/geo/query/acc.cgi?acc=GSE42861}{\(\scriptstyle\nearrow\)} \\
\nolinkurl{FusionAI-NTv2-GeneFusion} & Gene Fusion Breakpoints (NT-v2) & Embed. & C & 51,859 & 2,048 & \href{https://doi.org/10.5281/zenodo.18713246}{\(\scriptstyle\nearrow\)} \\
\nolinkurl{GEO-GSE50660-Methylation-Smoking} & Smoking Status DNA Methylation (GSE50660) & Meth. & C & 464 & 30,000 & \href{https://www.ncbi.nlm.nih.gov/geo/query/acc.cgi?acc=GSE50660}{\(\scriptstyle\nearrow\)} \\
\nolinkurl{GEO-GSE147221-Methylation-Schizophrenia} & Schizophrenia DNA Methylation (GSE147221) & Meth. & C & 679 & 30,000 & \href{https://www.ncbi.nlm.nih.gov/geo/query/acc.cgi?acc=GSE147221}{\(\scriptstyle\nearrow\)} \\
\nolinkurl{OpenML-46983} & Riboflavin Production & Expr. & R & 71 & 4,088 & \href{https://www.openml.org/search?type=data&sort=runs&id=46983&status=active}{\(\scriptstyle\nearrow\)} \\
\nolinkurl{OpenML-430} & mtp2 & Other & R & 274 & 1,142 & \href{https://www.openml.org/search?type=data&sort=runs&id=430&status=active}{\(\scriptstyle\nearrow\)} \\
\nolinkurl{TDC-AqSolDB} & Aqueous Solubility (AqSolDB) & Mol. & R & 9,562 & 2,048 & \href{https://dataverse.harvard.edu/api/access/datafile/4259610}{\(\scriptstyle\nearrow\)} \\
\nolinkurl{TDC-Lipophilicity-AstraZeneca} & Lipophilicity (AstraZeneca) & Mol. & R & 4,200 & 2,048 & \href{https://dataverse.harvard.edu/api/access/datafile/4259595}{\(\scriptstyle\nearrow\)} \\
\nolinkurl{TDC-LD50-Zhu} & Acute Toxicity LD50 (Zhu) & Mol. & R & 7,341 & 2,048 & \href{https://dataverse.harvard.edu/api/access/datafile/4267146}{\(\scriptstyle\nearrow\)} \\
\nolinkurl{GEO-GSE40279-Methylation-Age} & Blood DNA Methylation Age (GSE40279) & Meth. & R & 656 & 30,000 & \href{https://www.ncbi.nlm.nih.gov/geo/query/acc.cgi?acc=GSE40279}{\(\scriptstyle\nearrow\)} \\
\nolinkurl{local-PEER-BetaLactamase-ESM2} & TEM-1 Activity (ESM-2) & Embed. & R & 5,198 & 1,280 & \hyperref[app:local-construction]{\(\scriptstyle\nearrow\)} \\
\nolinkurl{gp-maize-FT} & Maize Flowering Time Genomic Prediction & SNP & R & 391 & 244,781 & \href{https://doi.org/10.5061/dryad.xksn02vb9}{\(\scriptstyle\nearrow\)} \\
\nolinkurl{gp-rice-FT} & Rice Flowering Time Genomic Prediction & SNP & R & 327 & 57,542 & \href{https://doi.org/10.5061/dryad.xksn02vb9}{\(\scriptstyle\nearrow\)} \\
\nolinkurl{gp-soy-YLD} & Soybean Yield Genomic Prediction & SNP & R & 5,014 & 4,234 & \href{https://doi.org/10.5061/dryad.xksn02vb9}{\(\scriptstyle\nearrow\)} \\
\nolinkurl{gp-switchgrass-HT} & Switchgrass Height Genomic Prediction & SNP & R & 514 & 217,150 & \href{https://doi.org/10.5061/dryad.xksn02vb9}{\(\scriptstyle\nearrow\)} \\
\nolinkurl{CHEMBL206} & Oestrogen receptor bioactivity & Mol. & R & 4,544 & 2,048 & \href{https://www.ebi.ac.uk/chembl/explore/target/CHEMBL206}{\(\scriptstyle\nearrow\)} \\
\nolinkurl{CHEMBL2034} & Glucocorticoid Receptor Bioactivity & Mol. & R & 3,127 & 2,048 & \href{https://www.ebi.ac.uk/chembl/explore/target/CHEMBL2034}{\(\scriptstyle\nearrow\)} \\
\end{longtable}

\endgroup
\newpage
\subsection{Target distributions}
\label{app:target-distributions}

Tables~\ref{tab:classification-targets} and~\ref{tab:regression-targets} describe the evaluated cohorts after filtering. They are reconstructed from the saved targets of the five disjoint held-out folds and match the cohort sizes in Table~\ref{tab:datasets}. Classification sampling is stratified rather than balanced to equal class sizes: budgeted training subsets retain at least one observation per class where possible, then allocate the remaining budget proportionally. These cohort summaries do not describe the allocation within each budgeted training subset. Defining roughly balanced as a largest class at most twice the size of the smallest, 14 of the 30 classification datasets (47\%) are roughly balanced and 16 (53\%) are imbalanced.

\begingroup
\fontsize{8}{8.5}\selectfont
\setlength{\tabcolsep}{3pt}
\renewcommand{\arraystretch}{0.92}
\begin{longtable}{@{}>{\raggedright\arraybackslash}p{0.28\linewidth}>{\raggedright\arraybackslash}p{0.13\linewidth}rr>{\raggedright\arraybackslash}p{0.43\linewidth}@{}}
\toprule
Registry ID & Target & $N$ & $K$ & Class: count (percentage) \\
\midrule
\endfirsthead
\toprule
Registry ID & Target & $N$ & $K$ & Class: count (percentage) \\
\midrule
\endhead
\midrule
\multicolumn{5}{r}{Continued on next page} \\
\endfoot
\bottomrule
\endlastfoot
\nolinkurl{OpenML-1138} & Tissue & 1,545 & 2 & Other: 1,421 (92.0\%); Uterus: 124 (8.0\%) \\[1pt]
\nolinkurl{OpenML-1458} & Class & 200 & 2 & 1: 88 (44.0\%); 2: 112 (56.0\%) \\[1pt]
\nolinkurl{TCGA-TCGA-BRCA_Gene-Expression-Quantification} & sample\_\allowbreak type & 1,231 & 2 & Normal: 113 (9.2\%); Tumor: 1,118 (90.8\%) \\[1pt]
\nolinkurl{GEO-GSE10893} & subtype & 165 & 5 & Basal: 40 (24.2\%); Her2: 21 (12.7\%); LumA: 52 (31.5\%); LumB: 40 (24.2\%); Normal-like: 12 (7.3\%) \\[1pt]
\nolinkurl{TCGA-TCGA-HNSC_Gene-Expression-Quantification} & sample\_\allowbreak type & 566 & 2 & Normal: 44 (7.8\%); Tumor: 522 (92.2\%) \\[1pt]
\nolinkurl{local-BRCA_Evo2-VEP} & y & 3,893 & 3 & FUNC: 2,821 (72.5\%); INT: 249 (6.4\%); LOF: 823 (21.1\%) \\[1pt]
\nolinkurl{local-DeepLoc2-Fungi-ESM2} & membrane & 5,841 & 2 & 0: 4,535 (77.6\%); 1: 1,306 (22.4\%) \\[1pt]
\nolinkurl{OpenML-1083} & Decision & 214 & 7 & 1: 13 (6.1\%); 2: 69 (32.2\%); 3: 21 (9.8\%); 4: 21 (9.8\%); 5: 35 (16.4\%); 6: 39 (18.2\%); 7: 16 (7.5\%) \\[1pt]
\nolinkurl{OpenML-1106} & class & 190 & 14 & Bladder: 11 (5.8\%); Breast: 11 (5.8\%); CNS: 20 (10.5\%); Colorectal: 11 (5.8\%); Leukemia: 30 (15.8\%); Lung: 11 (5.8\%); Lymphoma: 22 (11.6\%); Melanoma: 10 (5.3\%); Mesothelioma: 11 (5.8\%); Ovary: 11 (5.8\%); Pancreas: 11 (5.8\%); Prostate: 10 (5.3\%); Renal: 11 (5.8\%); Uterus\_\_Adeno: 10 (5.3\%) \\[1pt]
\nolinkurl{OpenML-1137} & Tissue & 546 & 2 & Colon: 286 (52.4\%); Kidney: 260 (47.6\%) \\[1pt]
\nolinkurl{OpenML-1084} & Decision & 220 & 3 & 1: 48 (21.8\%); 2: 44 (20.0\%); 3: 128 (58.2\%) \\[1pt]
\nolinkurl{OpenML-1088} & Decision & 383 & 9 & 2: 155 (40.5\%); 3: 24 (6.3\%); 4: 39 (10.2\%); 5: 47 (12.3\%); 6: 42 (11.0\%); 7: 20 (5.2\%); 8: 16 (4.2\%); 9: 18 (4.7\%); 10: 22 (5.7\%) \\[1pt]
\nolinkurl{OpenML-1140} & Tissue & 324 & 2 & Lung: 126 (38.9\%); Ovary: 198 (61.1\%) \\[1pt]
\nolinkurl{MGYS00005384} & host sex & 674 & 2 & female: 340 (50.4\%); male: 334 (49.6\%) \\[1pt]
\nolinkurl{MGYS00005380} & isolation source & 264 & 2 & gut: 91 (34.5\%); oral: 173 (65.5\%) \\[1pt]
\nolinkurl{MGYS00005285} & host sex & 438 & 2 & female: 179 (40.9\%); male: 259 (59.1\%) \\[1pt]
\nolinkurl{MGYS00006282} & host sex & 256 & 2 & female: 127 (49.6\%); male: 129 (50.4\%) \\[1pt]
\nolinkurl{MGYS00002425} & host sex & 394 & 2 & female: 236 (59.9\%); male: 158 (40.1\%) \\[1pt]
\nolinkurl{MGYS00005381} & environment (biome) & 218 & 2 & periurban shantytown: 70 (32.1\%); rural village: 148 (67.9\%) \\[1pt]
\nolinkurl{MGYS00001255} & host sex & 152 & 2 & female: 30 (19.7\%); male: 122 (80.3\%) \\[1pt]
\nolinkurl{MGYS00001056} & isolation source & 306 & 3 & gut: 94 (30.7\%); oral: 182 (59.5\%); vaginal: 30 (9.8\%) \\[1pt]
\nolinkurl{MGYS00005379} & environment (material) & 1,450 & 3 & cut: 63 (4.3\%); ss: 1,289 (88.9\%); sw: 98 (6.8\%) \\[1pt]
\nolinkurl{gut-cirrhosis} & disease & 232 & 2 & disease: 118 (50.9\%); healthy: 114 (49.1\%) \\[1pt]
\nolinkurl{TDC-BBB-Martins} & Y & 1,954 & 2 & 0: 461 (23.6\%); 1: 1,493 (76.4\%) \\[1pt]
\nolinkurl{TDC-AMES} & Y & 7,255 & 2 & 0: 3,304 (45.5\%); 1: 3,951 (54.5\%) \\[1pt]
\nolinkurl{TDC-CYP3A4-Veith} & Y & 12,300 & 2 & 0: 7,194 (58.5\%); 1: 5,106 (41.5\%) \\[1pt]
\nolinkurl{GEO-GSE42861-Methylation-RA} & disease state & 689 & 2 & Normal: 335 (48.6\%); rheumatoid arthritis: 354 (51.4\%) \\[1pt]
\nolinkurl{FusionAI-NTv2-GeneFusion} & fusion & 51,859 & 2 & 0: 25,903 (49.9\%); 1: 25,956 (50.1\%) \\[1pt]
\nolinkurl{GEO-GSE50660-Methylation-Smoking} & smoking status & 464 & 3 & current: 22 (4.7\%); former: 263 (56.7\%); never: 179 (38.6\%) \\[1pt]
\nolinkurl{GEO-GSE147221-Methylation-Schizophrenia} & status & 679 & 2 & case: 348 (51.3\%); control: 331 (48.7\%) \\[1pt]
\addlinespace[8pt]
\caption{Class distributions of the 30 classification datasets after filtering classes with fewer than ten observations. $N$ is the evaluated cohort size and $K$ the number of retained classes. Counts are pooled across the five disjoint held-out folds, so each observation is counted once. Labels are shown as stored, including numeric codes. Percentages are rounded.}
\label{tab:classification-targets} \\
\end{longtable}

\endgroup
\newpage
\begingroup
\fontsize{8}{8.5}\selectfont
\setlength{\tabcolsep}{3pt}
\renewcommand{\arraystretch}{0.92}
\begin{longtable}{@{}>{\raggedright\arraybackslash}p{0.25\linewidth}>{\raggedright\arraybackslash}p{0.14\linewidth}rrll@{}}
\toprule
Registry ID & Target & $N$ & Median & Interquartile interval & Range \\
\midrule
\endfirsthead
\toprule
Registry ID & Target & $N$ & Median & Interquartile interval & Range \\
\midrule
\endhead
\midrule
\multicolumn{6}{r}{Continued on next page} \\
\endfoot
\bottomrule
\endlastfoot
\nolinkurl{OpenML-46983} & x & 71 & -6.948 & $[-7.688,\, -6.449]$ & $[-9.966,\, -5.673]$ \\[1pt]
\nolinkurl{OpenML-430} & oz1143 & 274 & 0.358 & $[0.243,\, 0.461]$ & $[0.000,\, 1.000]$ \\[1pt]
\nolinkurl{TDC-AqSolDB} & Y & 9,562 & -2.633 & $[-4.320,\, -1.270]$ & $[-13.172,\, 2.138]$ \\[1pt]
\nolinkurl{TDC-Lipophilicity-AstraZeneca} & Y & 4,200 & 2.360 & $[1.410,\, 3.100]$ & $[-1.500,\, 4.500]$ \\[1pt]
\nolinkurl{TDC-LD50-Zhu} & Y & 7,341 & 2.367 & $[1.854,\, 3.034]$ & $[-0.343,\, 10.207]$ \\[1pt]
\nolinkurl{GEO-GSE40279-Methylation-Age} & age (y) & 656 & 65.000 & $[54.000,\, 75.000]$ & $[19.000,\, 101.000]$ \\[1pt]
\nolinkurl{local-PEER-BetaLactamase-ESM2} & activity & 5,198 & 0.877 & $[0.566,\, 0.982]$ & $[-0.008,\, 1.142]$ \\[1pt]
\nolinkurl{gp-maize-FT} & FT & 391 & -0.018 & $[-0.633,\, 0.553]$ & $[-2.000,\, 2.826]$ \\[1pt]
\nolinkurl{gp-rice-FT} & FT & 327 & 0.047 & $[-0.696,\, 0.729]$ & $[-2.684,\, 3.540]$ \\[1pt]
\nolinkurl{gp-soy-YLD} & YLD & 5,014 & 0.128 & $[-0.570,\, 0.705]$ & $[-4.283,\, 3.054]$ \\[1pt]
\nolinkurl{gp-switchgrass-HT} & HT & 514 & -0.014 & $[-0.712,\, 0.604]$ & $[-2.697,\, 3.006]$ \\[1pt]
\nolinkurl{CHEMBL206} & pchembl\_\allowbreak value & 4,544 & 7.000 & $[5.895,\, 8.192]$ & $[3.150,\, 11.000]$ \\[1pt]
\nolinkurl{CHEMBL2034} & pchembl\_\allowbreak value & 3,127 & 7.350 & $[6.560,\, 8.000]$ & $[4.000,\, 10.520]$ \\[1pt]
\addlinespace[8pt]
\caption{Target distributions of the 13 regression datasets, pooled across the five disjoint held-out folds. Values use the evaluated target scale; the four crop genomic-prediction targets are standardised. The interquartile interval is the 25th--75th percentile interval; the range gives the observed minimum and maximum. Values are rounded to three decimal places.}
\label{tab:regression-targets} \\
\end{longtable}

\endgroup
\newpage
\subsection{Construction of local embedding datasets}
\label{app:local-construction}

We constructed three sequence-representation datasets and loaded the resulting numerical tables through the same registry interface as the remotely retrieved datasets. The target and biological group columns were removed from the feature matrix and retained separately for evaluation and grouped cross-validation.

For the BRCA1 variant-effect task, we used the functional measurements of \citet{findlay2018}, also used in the Evo 2 BRCA1 evaluation \citep{evo2}. Our prepared input CSV contains paired reference and variant DNA sequences and the three-level functional label (FUNC, LOF, or INT). Each sequence was centre-cropped to the 8,192-base-pair context used in this benchmark and embedded with Evo2-7B at layer \texttt{blocks.26.mlp.l3}. We mean-pooled the token-level representation of each sequence and concatenated its 4,096-dimensional reference and variant vectors to form an 8,192-dimensional feature vector. A truncated SHA-256 hash of the reference sequence identified variants from the same genomic context and served as the cross-validation group.

For the protein tasks, source files were downloaded from their official hosts and verified against pinned checksums. We embedded proteins with the pinned \texttt{facebook/esm2\_t33\_650M\_UR50D} revision \citep{esm2} and mean-pooled its 1,280-dimensional residue representations. Proteins longer than the model context were split into non-overlapping 1,022-residue chunks and pooled using residue-count weighting, so every residue contributed once. The DeepLoc 2.0 dataset contains all 5,841 unique fungal proteins in the official Swiss-Prot training and validation table; membrane association is the binary target, and the five published homology partitions define the cross-validation groups \citep{deeploc2}. The PEER TEM-1 dataset combines the official training, validation, and test files into 5,198 unique single-mutant sequences; the scaled experimental activity is the regression target, and the position of the sole amino-acid substitution defines 286 cross-validation groups \citep{peer}. The generated protein tables were validated for finite, non-constant features and accompanied by manifests that record the model revision, source and output digests, pooling procedure, and dimensions.



\end{document}